\documentclass[10pt]{article} 
\usepackage[accepted]{tmlr}

\usepackage{amsmath,amsfonts,bm}

\def\eqref#1{equation~\ref{#1}}

\def\1{\bm{1}}

\DeclareMathAlphabet{\mathsfit}{\encodingdefault}{\sfdefault}{m}{sl}
\SetMathAlphabet{\mathsfit}{bold}{\encodingdefault}{\sfdefault}{bx}{n}

\usepackage{hyperref}
\usepackage{url}
\usepackage{graphicx}
\usepackage{algorithm}
\usepackage{algpseudocode} 
\usepackage[table]{xcolor}
\usepackage{xcolor}  
\newcommand{\update}[1]{\textcolor{black}{#1}}

\title{Don’t Let It Hallucinate: Premise Verification via \\ Retrieval-Augmented Logical Reasoning}

\author{\name Yuehan Qin \email yuehanqi@usc.edu \\
      \addr University of Southern California
      \AND
      \name Shawn Li \email li.li02@usc.edu \\
      \addr University of Southern California
      \AND
      \name Yi Nian \email yinian@usc.edu\\
      \addr University of Southern California \\
      \AND
      \name Xinyan Velocity Yu \email xinyany@usc.edu\\
      \addr University of Southern California \\
      \AND
      \name Yue Zhao\thanks{Corresponding authors.} \email yue.z@usc.edu\\
      \addr University of Southern California \\
      \AND
      \name Xuezhe Ma\footnotemark[1] \email xuezhema@usc.edu\\
      \addr University of Southern California \\
      }

\def\month{02}  
\def\year{2026} 
\def\openreview{\url{https://openreview.net/forum?id=BDxStRGWba}} 

\begin{document}

\maketitle

\begin{abstract}
Large language models (LLMs) have shown substantial capacity for generating fluent, contextually appropriate responses. 
However, they can produce hallucinated outputs, especially when a user query includes one or more \emph{false premises}—claims that contradict established facts. Such premises can mislead LLMs into offering fabricated or misleading details. 
Existing approaches include pretraining, fine-tuning, and inference-time techniques that often rely on access to logits or address hallucinations after they occur. 
These methods tend to be computationally expensive, require extensive training data, or lack proactive mechanisms to prevent hallucination before generation, limiting their efficiency in real-time applications.
We propose a retrieval-based framework that identifies and addresses false premises \emph{before} generation.
Our method first transforms a user’s query into a logical representation, then applies retrieval-augmented generation (RAG) to assess the validity of each premise using factual sources. 
Finally, we incorporate the verification results into the LLM’s prompt to maintain factual consistency in the final output. 
Experiments show that this approach effectively reduces hallucinations, improves factual accuracy, and does not require access to model logits or large-scale fine-tuning.
\end{abstract}

\section{\update{Introduction}}
Large Language Models (LLMs) generate fluid, context-aware responses but can produce hallucinations when prompted with queries that include hidden factual errors \citep{manakul2023selfcheckgptzeroresourceblackboxhallucination, zheng2023judgingllmasajudgemtbenchchatbot}. 
These errors, known as \emph{false premises}, are statements in a user’s question that conflict with real-world facts. Even when LLMs can store accurate information, they may trust the incorrect assumptions embedded in the query and generate misleading outputs \citep{yuan2024whispersshakefoundationsanalyzing}. 
This is especially problematic in sensitive applications such as finance or healthcare, where mistakes can cause serious harm \citep{pal2023medhaltmedicaldomainhallucination}.

\begin{figure*}[h]
  \centering
  \includegraphics[width=\textwidth]{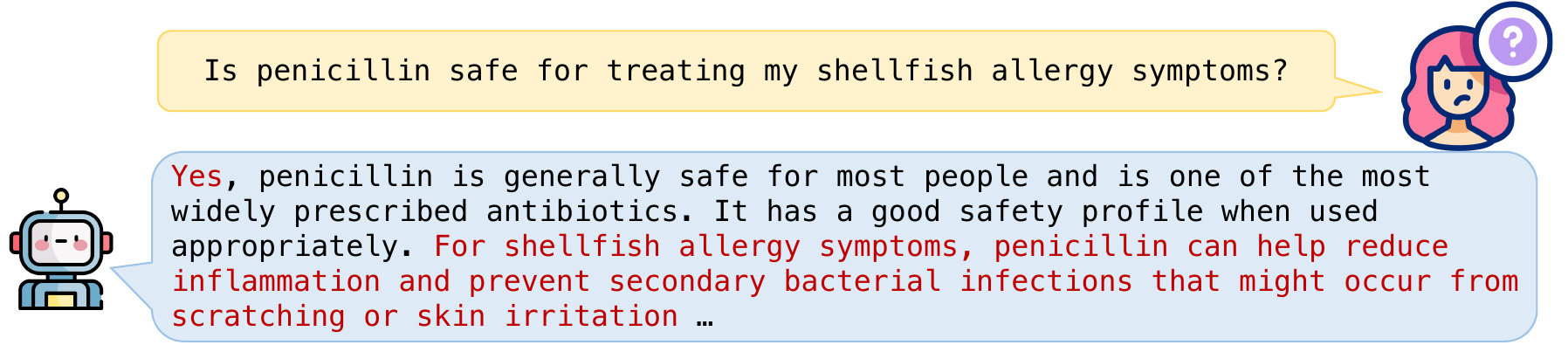}
  \caption{LLM experiences factuality hallucination when faced with a false premise question, where both entities 
  \textit{shellfish allergy symptom} and \textit{penicillin}
  exist but are not correctly aligned. The LLM's hallucinated response could delay life-saving treatment by incorrectly recommending antibiotics for allergic reactions.}
  \label{fig:f2}
\end{figure*}

Prior research distinguishes between \emph{factuality} hallucinations, where the output conflicts with known facts, and \emph{faithfulness} hallucinations, where the response diverges from the provided context or user instructions \citep{10.1145/3637528.3671796}. 
We focus on factuality hallucinations, particularly those driven by incorrect assumptions (false premises). These errors are common among LLM outputs \citep{Huang_2025, snyder2024early}, where Fig.~\ref{fig:f2} presents an example of a question with a false premise and the resulting hallucination. These premise-driven errors are particularly insidious as they can appear factually sound while being fundamentally incorrect.

Many methods attempt to address false premises after an LLM has already produced an answer. 
They include fine-tuning the model to detect invalid assumptions \citep{hu2023wontfooledagainanswering}, applying contrastive decoding to surface inconsistencies \citep{shi2023trustingevidencehallucinatecontextaware, chuang2024doladecodingcontrastinglayers}, and using uncertainty-based measures or logits to gauge inaccuracies \citep{pezeshkpour2023measuringmodifyingfactualknowledge,varshney2023stitchtimesavesnine}. 
Although effective in some contexts, these approaches can be computationally demanding and do not necessarily prevent misinformation from appearing in the first place. 
Additionally, questions with false premises often maintain normal semantic flow, changing only a few tokens so that they are difficult to identify using traditional out-of-distribution detection \citep{vu2023freshllmsrefreshinglargelanguage}. 
Even advanced LLMs can struggle with real-time truth evaluation, lacking the context or capacity to fully check every assumption \citep{hu2023wontfooledagainanswering, liu2024selfcontradictoryreasoningevaluationdetection}.

\begin{figure*}[t]
  \centering
  \includegraphics[width=\textwidth]{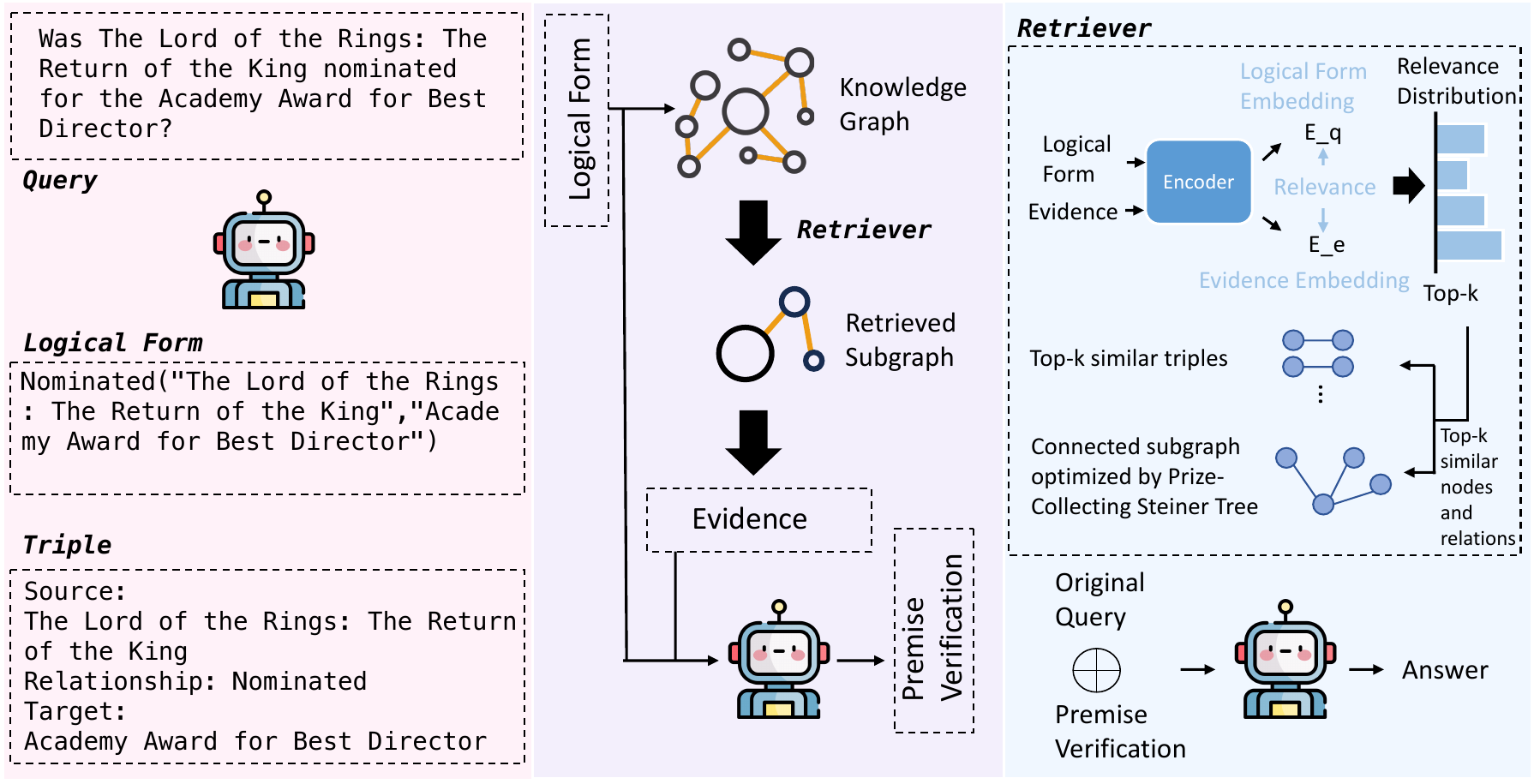}
  \caption{Overview of our approach. Left: The original query is converted into a logical form. Middle: The logical form is used to retrieve relevant elements from the knowledge graph and detect false premises. Right: Comparison of studied retrievers for aligning logical form with the knowledge graph. The LLM generates responses with reduced hallucination given prompts with premise verification.}
  \label{fig:f1}
\end{figure*}

To address this challenge, we focus on \emph{preventing} hallucinations rather than mitigating them post hoc. In our framework, 
we first transform the user’s query into a logical form that highlights key entities or relations.
We then employ retrieval-augmented generation (RAG) to check the accuracy of these statements against a knowledge graph.
If contradictions are found, the query is flagged as containing a false premise
prompting the model to correct or reject the assumption before formulating a final answer. 
This process, shown in Fig.~\ref{fig:f1}, ensures that the LLM does not rely on erroneous details during response generation. By informing the LLM about any detected false premise in advance,
we reduce the likelihood of hallucinations without requiring access to model logits or large-scale fine-tuning.

We summarize our contributions as follows:

\textbf{Logical Form Representation}: We first introduce logical forms to represent input queries and demonstrate their effectiveness across various types of graph retrievers.
This logical approach enables accurate and systematic evaluation of statements provided in user prompts, particularly handling queries that may include false premises.

\textbf{Explicit False Premise Detection}:
Our method improves the reliability of LLM-generated responses by explicitly detecting false premises and informing the LLM if a question contains a false premise.

\textbf{Hallucination Mitigation Without Output Generation or Model Logits}: Our approach reduces factual hallucinations without actual generation of responses or LLM logits and, therefore, can be seamlessly integrated into existing LLM frameworks and pipelines, offering a straightforward enhancement for improving factual accuracy.

\section{Related Works}
\textbf{False Premise}. A False Premise Question (FPQ) is a question containing incorrect facts that are not necessarily explicitly stated but might be mistakenly believed by the questioner \citep{yu2022crepeopendomainquestionanswering, kim2021linguistinventedlightbulbpresupposition}. Recent studies \citep{yuan2024whispersshakefoundationsanalyzing,Li_Ji_Wu_Li_Qin_Wei_Zimmermann_2024,Li_2025_CVPR} have demonstrated that FPQs can induce factuality hallucination in LLMs, as they often respond directly to FPQs without verifying their validity.
Notably, existing prompting techniques like few-shot prompting \citep{brown2020languagemodelsfewshotlearners} and Chain-of-Thought \citep{wei2023chainofthoughtpromptingelicitsreasoning}, tend to increase hallucinations.
Conversely, directly prompting LLMs to detect false premises degrades their performance on questions containing valid premises \citep{vu2023freshllmsrefreshinglargelanguage}.

\textbf{Logical Forms}. Symbolic solvers and logical forms are applied to logical reasoning by grounding natural language in symbolic representations. The latest trend is integrating LLMs with symbolic solvers to enhance their performance \cite{Olausson_2023, pan2023logiclmempoweringlargelanguage,li2025secureondevicevideoood,limm}, where natural language is translated into symbolic logic forms and deterministic symbolic solvers are employed for inference, enabling more accurate logical problem-solving. Similarly, 
SymbCoT \cite{xu2024faithfullogicalreasoningsymbolic}
converts input text into symbolic formats such as first-order logic, generates reasoning plans through logical rule application, and verifies the reasoning process to ensure consistency. These methods demonstrate that incorporating symbolic improves the reliability and interpretability of LLM outputs, making them well-suited for tasks requiring logical consistency.

\textbf{Knowledge Graph Fact Checking and Question Answering}. In fact checking, RAG approaches verify data accuracy, with knowledge graph-driven RAG gaining attention for effectively leveraging structured knowledge. Recent works include: 1) \textit{prompt-based} methods where \citep{pan2023qacheckdemonstrationquestionguidedmultihop} evaluates evidence sufficiency and generates verification questions, and \citep{sun2024thinkongraphdeepresponsiblereasoning} performs hop-by-hop fact retrieval; 2) \textit{graph-based} approaches where \citep{he2024gretrieverretrievalaugmentedgenerationtextual} formulates RAG as a Prize-Collecting Steiner Tree problem for subgraph extraction, and \citep{mavromatis2024gnnraggraphneuralretrieval} uses graph neural networks for dense subgraph reasoning and answer retrieval;
3) \textit{training-based} methods where \citep{zheng-etal-2024-evidence} develops dual encoders for query and subgraph evidence embedding, and \citep{liu2024knowledgegraphenhancedlargelanguage} trains encoders for retrieval and ranking processes, though requiring entity presence in the knowledge graph and relying on prompt-generated training data.

\textbf{Hallucination Mitigation}. 
Sources of LLM hallucinations originate from different stages in the LLM life cycle \citep{zhang2023languagemodelhallucinationssnowball,li2026defensespromptattackslearn,li-etal-2025-treble}, leading existing mitigation methods to target specific stages:
1) \textit{Pre-training}: Enhancing factual reliability by emphasizing credible texts, either by up-sampling trustworthy documents \citep{touvron2023llama2openfoundation} or prepending factual sentences with topic prefixes \citep{lee2023factualityenhancedlanguagemodels}.
 2) \textit{Supervised Fine-tuning}: Curating high-quality, instruction-oriented datasets \citep{chen2024alpagasustrainingbetteralpaca,cao2024instructionmininginstructiondata}
 improves factual accuracy more effectively than fine-tuning on unfiltered data, and remains more feasible compared to extensive pre-training.
 3) \textit{Reinforcement Learning from Human Feedback}: Aligning closely with human preferences may inadvertently encourage hallucinations or biased outputs, especially when instructions surpass the model’s existing knowledge \citep{radhakrishnan2023questiondecompositionimprovesfaithfulness, wei2024simplesyntheticdatareduces}. 
 4) \textit{Inference}: Known as hallucination snowballing \citep{zhang2023sirenssongaiocean}, LLMs occasionally magnify initial generation mistakes. Proposed inference-time solutions include new decoding strategies \citep{shi2023trustingevidencehallucinatecontextaware, chuang2024doladecodingcontrastinglayers}, uncertainty analysis of model outputs \citep{ xu2025decopromptdecodingprompts, liu2024dellmadecisionmakinguncertainty,dhuliawala2023chainofverificationreduceshallucinationlarge,li2025personalizedconversationalbenchmarksimulating}. However, these approaches either act post-hallucination or require access to model logits, thus being inefficient due to repeated prompting or limited to white-box LLM scenarios.
 We briefly discuss the comparison between our work and previous post-hoc hallucination mitigation method in Tab.~\ref{tab:training_comparison} and detailed discussion can be found in Discussion and Appendix \S~\ref{add_result}.

\begin{table*}[h]
\centering
\small
\scalebox{0.88}{
\begin{tabular}{lccccc}
\hline
\textbf{Method} & \textbf{Training cost} & \textbf{Number of tokens} & \textbf{Training time} & \textbf{Model agnostic} & \textbf{Black-box Compatible} \\
\hline
\rowcolor{gray!20}
Post-hoc & Depends on& Train: original query + answer & Depends on& No & No \\
\rowcolor{gray!20}
method & fine-tuning & Inference: original query &  fine-tuning  & & \\
Ours & Zero & Original query + logical form & Zero & Yes & Yes \\
\hline
\end{tabular}}
\caption{Comparison of training and compatibility between post-hoc method and our method.}
\label{tab:training_comparison}
\end{table*}

\section{Methodology}
\label{sec:method}
LLM hallucinations often stem from false premises in user queries. Instead of addressing hallucinations post-generation, we aim to prevent them by detecting 
and informing the presence of false premises to LLMs before response generation. Our proposed method achieves this through three key steps:

\textbf{Logical Form Conversion}.
By converting the user query into a structured logical form representation, we extract its core meaning, making it easier to analyze its factual consistency. We demonstrate its effectiveness across various types of graph retrievers.

\textbf{Structured Retrieval and Verification}. 
Rather than relying solely on model-generated text, we retrieve external evidence to assess whether false premises exist in the query.

\textbf{Factual Consistency Enforcement}.
The verified information is then incorporated into the LLM prompt, ensuring that the model generates responses aligned with factual data.

Our proposed method applies to knowledge graphs and datasets compatible with graph structures.
We show the pseudocode summary of our approach in Algorithm~\ref{alg:alg}.
\begin{algorithm}[h]
\footnotesize 
\caption{False premise detection and hallucination mitigation}
\label{alg:alg}
\renewcommand{\algorithmicrequire}{\textbf{Input:} 
 User query $q$, Knowledge graph $G$
}
\renewcommand{\algorithmicensure}{\textbf{Output:} 
 Hallucination mitigated response from LLM
}
\begin{algorithmic}[1]
\Require $ $
\Ensure $ $

\State Convert user query $q$ into logical representation $\mathcal{L}(q)$ \Comment{(\S \ref{sec:logical_form})}
\State Extract logical assertions $P(x_1, x_2, \dots, x_n)$ from $\mathcal{L}(q)$

\State Initialize maximum similarity score $Sim_{max} \gets -\infty$ \Comment{(\S \ref{sec:retrieval})}
\State Initialize optimal graph $G^* \gets \emptyset$
\State Candidate set $G^* \gets$ subsets of relevant subgraphs from $G$, i.e., $R(G)$

\For {triple $G' \in G$}
    \If{retriever is embedding-based}
        \State Compute similarity via embeddings:
        $$Sim \gets \text{Sim}\left(\mathcal{L}(q), G'\right)$$
    
    \ElsIf{retriever is non-parametric}
        \State Compute similarity using tree search criteria:
        $$Sim \gets \text{PCST}\left(\mathcal{L}(q), G'\right)$$
    
    \ElsIf{retriever is LLM-based}
        \State Compute similarity using LLM scoring:
        $$Sim \gets \text{LLMScore}\left(\mathcal{L}(q), G'\right)$$
    
    \EndIf
    \If{$Sim > Sim_{max}$}
        \State $Sim_{max} \gets Sim$
        \State $G^* \gets G'$
    \EndIf
\EndFor

\State Define false premise indicator function: \Comment{(\S \ref{sec:false_premise_detection})}
\[
F(q)=
\begin{cases}
1, & \text{if $q$ conflicts with retrieved evidence } G^*=R(q,G^*)\\
0, & \text{otherwise}
\end{cases}
\]

\If{$F(q)=1$} \Comment{(\S \ref{sec:hallucination_mitigation})}
    \State Update query as:
    \[
    q \gets q + \text{" Note: This question contains a false premise."}
    \]
\EndIf
\State Generate response from LLM using updated query $q$
\State \Return Hallucination mitigated response from LLM
\end{algorithmic}
\end{algorithm}

\subsection{Problem Definition}\label{sec:false_premise_detection}
\textbf{False Premise Detection}: Given a user query \( q \), the function \( F(q) \) determining whether \( q \) contains a false premise can be defined as:
\begin{equation} \label{problem_dfn}
F(q) =
\begin{cases} 
1, & \text{if } q \text{ conflicts with retrieved evidence } R(q, G), \\
0, & \text{otherwise},
\end{cases}
\end{equation}
where \( R\) denotes the retrieval function that extracts relevant evidence from a knowledge graph \( G \). The query \( q \) is evaluated against \( R(q, G) \), and if contradictions are found, \( q \) is deemed to contain a false premise (\( F(q) = 1 \)); otherwise, it is considered valid (\( F(q) = 0 \)).
In this study, the function $F$ is achieved by RAG using a retriever that leverages logical form and a knowledge graph.

\subsection{Logical Form Extraction}\label{sec:logical_form}
\textbf{Logical Form}: A logical form is a structured representation of statements or queries expressed using symbolic logic. It provides a structured way to capture semantic relationships within sentences, enabling precise and systematic reasoning. Given a natural language sentence query $q$, its logical form $\mathcal{L}(q)$ can be represented as:
$\mathcal{L}(q) = P(x_1, x_2, \dots, x_n),$
where $P$ denotes a predicate or relation, and $x_1, x_2, \dots, x_n$ are variables or constants representing entities or concepts extracted from $q$. For example, for the query:

\textit{Was The Lord of the Rings: The Return of the King nominated for the Academy Award for Best Director?}

Its logical form is:

\textit{Nominated("The Lord of the Rings: The Return of the King", "Academy Award for Best Director")}

Here, \textit{"The Lord of the Rings: The Return of the King"} and \textit{"Academy Award for Best Director"} are entities, while \textit{Nominated} is the relation.
GPT-4o-mini \citep{openai2024gpt4technicalreport}  is used for extracting logical forms from the queries. For an input query $q$, we first ask the LLM to generate the corresponding logical form $L(q)$. Then, we extract the source, relationship, and target from $L(q)$.
The prompt for logical form conversion is included in Appendix \S~\ref{sec:prompt}.

To evaluate the quality of the logical form conversion, we ask two annotators to manually grade the generated logical forms using a three-point scale: 1 (do not match), 2 (partially match), and 3 (match). Each annotator grades 100 randomly sampled outputs. Across all 200 samples, the generated logical forms receive a score of 3 in all of the cases. \update{More details are in Appendix \S \ref{sec:annotator}}.

\subsection{Retrieval}\label{sec:retrieval}
Given a user query $q$ in natural language, the
retrieval stage aims to extract the most relevant elements (e.g., entities, triplets, paths, subgraphs) from knowledge graphs, which can be formulated as:
\begin{align}\label{algo:eq3}
G^* 
&= \text{Graph-Retriever}(q, G) \nonumber \\
&= \arg\max_{G \subseteq R(G)} p_{\theta}(G \mid q, G) \nonumber \\
&= \arg\max_{G \subseteq R(G)} \text{Sim}(q, G),
\end{align}
where $G^*$ is the optimal retrieved graph elements, and $\text{Sim}(\cdot, \cdot)$ is a function that measures the semantic similarity between user queries and the graph data. $R(\cdot)$ represents a function to narrow down the search range of subgraphs, considering the efficiency.

After converting a user query $q$ into a logical form representation $L(q)$, the retriever encodes the logical form and the graph triples, searches through the knowledge graph $G$, and extracts the most relevant triple or subgraph, applying different selection criteria depending on the retriever used in our study.
Therefore, formula~\ref{algo:eq3} can be further formulated to: 
\begin{align}\label{algo:eq4}
G^* 
&= \text{Graph-Retriever}(\mathcal{L}(q), G) \nonumber \\
&= \arg\max_{G \subseteq R(G)} p_{\theta}(G \mid \mathcal{L}(q), G) \nonumber \\
&= \arg\max_{G \subseteq R(G)} \text{Sim}(\mathcal{L}(q), G).
\end{align}
We employ the pre-trained encoder \textit{all-roberta-large-v1}\footnote{https://huggingface.co/sentence-transformers/all-roberta-large-v1} to encode the logical form and graph triplets. The representation $L(q)$ is used in both the similarity-based retrieval process and the step where the LLM assesses whether the original query $q$ contains a false premise.

\subsection{Hallucination Mitigation}\label{sec:hallucination_mitigation}
For a given user query \( q \), if the false premise identification function \( F(q) \) detects a false premise (\( F(q) = 1 \)), we update its original query $q$ by appending a note:
\begin{equation}
q =
\begin{cases}
q + W, & \text{if } F(q) = 1, \\
q, & \text{otherwise},
\end{cases}
\end{equation}
where $W$ = "Note: This question 
contains a false premise.", and $q$ is the modified query that explicitly flags the presence of a false premise when detected. Once the original query is updated, we evaluate LLM's responses and measure the effectiveness of the ensuing hallucination mitigation.

\section{Experiments}
\subsection{Dataset}
KG-FPQ \citep{zhu2024kgfpqevaluatingfactualityhallucination} is a dataset containing true and false premise questions that are constructed from the KoPL knowledge graph, a high-quality subset of Wikidata. TPQs are generated from true triplets, while FPQs are created by replacing objects in false triplets via string matching. We evaluate the discriminative task in the art domain, where LLMs answer Yes-No questions (e.g., Is Hercules a cast member of 'The Lord of the Rings: the Return of the King'?). Dataset details are in Appendix \S \ref{sec:kgdataset}.

The CREAK dataset \citep{onoe2021creakdatasetcommonsensereasoning} is a benchmark for commonsense reasoning about entity knowledge. Unlike prior datasets focused on general physical or social scenarios, CREAK targets inferences that combine factual knowledge about specific entities with commonsense reasoning. It contains 13,000 human-authored English claims about entities labeled as true or false, along with a small contrast set. Each example requires understanding both factual attributes and commonsense implications. Dataset details are in Appendix \S \ref{sec:creakdataset}.

\update{The FEVER dataset \cite{thorne-etal-2018-fever} consists of natural-language claims labeled as \textit{Supported}, \textit{Refuted}, or \textit{Not Enough Information}. Evidence is retrieved from Wikipedia, which functions as a natural-language knowledge base. Claims are created by modifying Wikipedia sentences and are subsequently verified independently, without access to their original sources. FEVER is widely adopted as a standard benchmark for fact verification and claim validation.}

\subsection{Experiment Setting}
Our approach mitigates hallucination through a two-step process: First, we detect false premises in the user query. 
Then, we use the result of false premise detection along with the original query when providing input to the LLM.
We use both the KG-FPQ dataset and the CREAK dataset for evaluating the premise detection task, and we use the KG-FPQ dataset for the hallucination mitigation task, since CREAK dataset contains statements, not questions.

\subsubsection{False Premise Detection with Logical Form}
In the false premise detection task, we look at different retrievers with and without the use of logical forms. Logical forms are used in 1) the retrieval stage, where the logical form $\mathcal{L}(q)$ is encoded to find the most relevant elements from knowledge graph $G$, and 2) the false premise detection stage, where the logical form is passed as input along with the retrieved evidence to LLM to determine whether the query contains false premise.
The prompt detail is in 
Appendix~\ref{sec:prompt}.
We evaluate the use of logical forms in three configurations: 1) applying logical forms in both the retrieval stage and false premise detection stage, 2) using logical forms for retrieval and employing the original query for false premise detection, and 3) utilizing the original query for both stages. \update{We further analyze the role of different components in the logical form through ablation studies in Appendix~\ref{app:logical_form_ablation}}.

\subsubsection{False Premise Detection Methods}\label{sec:4_2}
We evaluate how logical form impacts retrieval for false premise detection across the following retrievers:

1) \textbf{Direct Claim}: We directly query the LLM to determine whether the given question contains a false premise. The model is prompted with:
\textit{Does the following question contain a false premise? Answer with 'Yes' or 'No' only.}

2) \textbf{Embedding-based Retriever}: \textbf{\textit{with RAG}} selects the top-k\footnote{This work focuses on top-1 selection.} relevant triples from the knowledge graph based on the cosine similarity between the query embedding and the graph triple embedding.

3) \textbf{Non-parametric Retriever}: \textbf{\textit{G-retriever}} \citep{he2024gretrieverretrievalaugmentedgenerationtextual} uses Prize-Collecting Steiner Tree algorithm for extracting relevant subgraph from the knowledge graph. It does not rely on a trained model with learnable parameters.

4) \textbf{LLM-based Retriever}: \textbf{\textit{GraphRAG/ToG}} \citep{edge2025localglobalgraphrag, sun2024thinkongraphdeepresponsiblereasoning} asks the LLM to generate a score between 0 and 100, indicating how helpful the generated answer is in answering the target question. The answers are sorted in descending order
of helpfulness score and used to generate the final answer returned to the user. 

5) \textbf{SAC3} \cite{zhang2024sac3reliablehallucinationdetection} is a hallucination detection approach that identifies hallucinations by assessing semantic-aware cross-check consistency, which involves generating semantically equivalent question perturbations and performing cross-model response consistency verification. We include SAC3 as baseline for the KG-FPQ dataset.

We use GPT-4o-mini
as the LLM in the false premise detection task. \update{Additional evaluations on other LLMs are provided in Appendix~\S \ref{sec:llama_detection}}.
These retrievers are included because they enable retrieval without task-specific fine-tuning, making them more adaptable across different domains. Unlike training-based retrievers, which require labeled data and extensive computation, non-parametric retriever uses structured knowledge, embedding-based retriever utilizes pre-trained encoders to transform queries and knowledge into a shared vector space for efficient retrieval, and LLM-based retrieval leverages pre-trained language models' generalization abilities. This setup evaluates the impact of logical forms on retrieval efficiency without the overhead of model training.

\textbf{Metrics.} 
We evaluate the false premise detection task using TPR (true positive rate), TNR (true negative rate), FPR (false positive rate), FNR (false negative rate), F1 score, and accuracy
of the model successfully identifying questions containing false premises or not. Here, a \textit{positive} instance refers to a question that contains a false premise. Higher TPR indicates better detection of false premises.

\begin{table*}[h]
\centering
\renewcommand{\arraystretch}{1}
\begin{tabular}{lcccc}
\hline
 & \textbf{Direct Claim} & \textbf{with RAG}  & \textbf{G-retriever} & \textbf{GraphRAG/ToG} \\ \hline

\multicolumn{5}{c}{Original Query for Both Stages} \\ \hline
True Positives (TP\%)  & 44.44  & 33.33  & 88.89 & 8.89  \\
True Negatives (TN\%)  & 73.33  & 80.00  &  86.67 & 93.33  \\
False Positives (FP\%)  & 26.67  & 20.00  &  13.33  & 6.67  \\
False Negatives (FN\%)  & 55.56  & 66.67  &  11.11 & 91.11  \\ 
F1 Score (\%)          & 59.70  & 48.78  &  87.89 & 16.16  \\
Accuracy (\%)          & 69.20 & 73.33 &  88.57 & 81.27 \\ \hline

\multicolumn{5}{c}{Logical Form for Retrieval and Original Query for False Premise Detection} \\ \hline
True Positives (TP\%)  & 44.44  & 37.78  &  82.22 & 8.89  \\
True Negatives (TN\%)  & 73.33 & 86.67  &  93.33 & 93.33  \\
False Positives (FP\%)  & 26.67  & 13.33  & 6.67 & 6.67  \\
False Negatives (FN\%)  & 55.56  & 62.22  &  17.78 & 91.11  \\ 
F1 Score (\%)          & 59.70  & 53.97  &  86.97 & 16.16  \\ 
Accuracy (\%)          & 69.20 & 79.69 &  83.81 & 81.27 \\ \hline

\multicolumn{5}{c}{Logical Form for Both Stages} \\ \hline
True Positives (TP\%)  & 44.44  & 60.00  &  94.44 & 8.89  \\
True Negatives (TN\%)  & 73.33  & 86.67  &  99.05 & 93.33  \\
False Positives (FP\%)  & 26.67  & 13.33  &  0.95 & 6.67  \\
False Negatives (FN\%)  & 55.56  & 40.00  &  5.56 & 91.11  \\ 
F1 Score (\%)          & 59.70  & \textbf{73.97}  &  \textbf{97.12} & 16.16  \\
Accuracy (\%)          & 69.20  & \textbf{82.86} & \textbf{95.24} & 81.27 \\ \hline
\end{tabular}
\caption{KG-FPQ dataset: comparison of performance metrics across different retrieval methods using logical forms and/or original queries.}
\label{tab:false_premise_detection_perf}
\end{table*}

\begin{table*}[h]
\centering
\renewcommand{\arraystretch}{1}
\begin{tabular}{lcccc}
\hline
 & \textbf{Direct Claim} & \textbf{with RAG}  & \textbf{G-retriever} & \textbf{GraphRAG/ToG} \\ \hline

\multicolumn{5}{c}{Original Query for Both Stages} \\ \hline
True Positives (TP\%)  & 72.5  & 62.3  & 24.6 & 89.9  \\
True Negatives (TN\%)  & 89.7  & 86.8  & 92.6 & 91.2  \\
False Positives (FP\%)  & 10.3  & 13.2  &  7.4  & 8.8  \\
False Negatives (FN\%)  & 27.5  & 37.7  &  75.4 & 10.1  \\ 
F1 Score (\%)          & 79.4  & 71.1 &  37.4 & 90.5  \\
Accuracy (\%)          & 81.0 & 74.5 & 58.4 & 90.5 \\ \hline

\multicolumn{5}{c}{Logical Form for Retrieval and Original Query for False Premise Detection} \\ \hline
True Positives (TP\%)  & 72.5  & 76.8  & 36.2 & 89.9  \\
True Negatives (TN\%)  & 89.7  & 92.6  & 83.8 & 88.2  \\
False Positives (FP\%)  & 10.3  & 7.4  &  16.2  & 11.8  \\
False Negatives (FN\%)  & 27.5  & 23.2  & 63.8 & 10.1  \\ 
F1 Score (\%)          & 79.4  & 83.5  &  47.6 & 89.2  \\
Accuracy (\%)          & 81.0 & 84.7 &  59.9 & 89.1 \\ \hline

\multicolumn{5}{c}{Logical Form for Both Stages} \\ \hline
True Positives (TP\%)  & 72.5  & 88.4  & 92.8 & 92.8  \\
True Negatives (TN\%)  & 89.7  & 92.6  & 83.8 & 91.2 \\
False Positives (FP\%)  & 10.3  & 7.4  &  16.2  & 8.8  \\
False Negatives (FN\%)  & 27.5  & 11.6  &  7.2 & 7.2  \\ 
F1 Score (\%)          & 79.4  & \textbf{90.4}  &  \textbf{88.9} & \textbf{92.1}  \\
Accuracy (\%)          & 81.0 & \textbf{90.5} &  \textbf{88.3} & \textbf{92.0} \\ \hline

\end{tabular}
\caption{CREAK dataset: comparison of performance metrics across different retrieval methods using logical forms and/or original queries.}
\label{tab:creak}
\end{table*}

\begin{table}[t]
\centering
\small
\setlength{\tabcolsep}{5pt}
\begin{tabular}{lccccc}
\hline
& \textbf{Direct} 
& \multicolumn{3}{c}{\textbf{WRAG (with RAG)}} 
& \textbf{GraphRAG/ToG} \\
\cline{3-5}
&  
& Orig
& LF-Retr 
& LF-Both 
&  \\
\hline
True Positive (TP\%)  & 18.8 & 96.1 & 94.1 & 94.1 & 88.2 \\
True Negative (TN\%)  & 98.5 & 61.2 & 83.7 & 83.7 & 83.7 \\
False Positive (FP\%) & 1.5  & 38.8 & 16.3 & 16.3 & 16.3 \\
False Negative (FN\%) & 81.2 & 3.9  & 5.9  & 5.9  & 11.8 \\
F1 Score (\%)         & 31.3 & 82.4 & 89.7 & 89.7 & 86.5 \\
Accuracy (\%)         & 58.4 & 79.0 & 89.0 & 89.0 & 86.0 \\
\hline
\end{tabular}
\caption{\update{FEVER dataset: comparison of performance metrics across different retrieval methods using logical
forms and/or original queries. Since g-retriever is a graph-based retriever and does not apply to non-graph data, we do not include it here.}}
\label{tab:fever}
\end{table}

\begin{figure*}[h]
    \small
  \centering
  \scalebox{0.9}{
  \includegraphics[width=\textwidth]{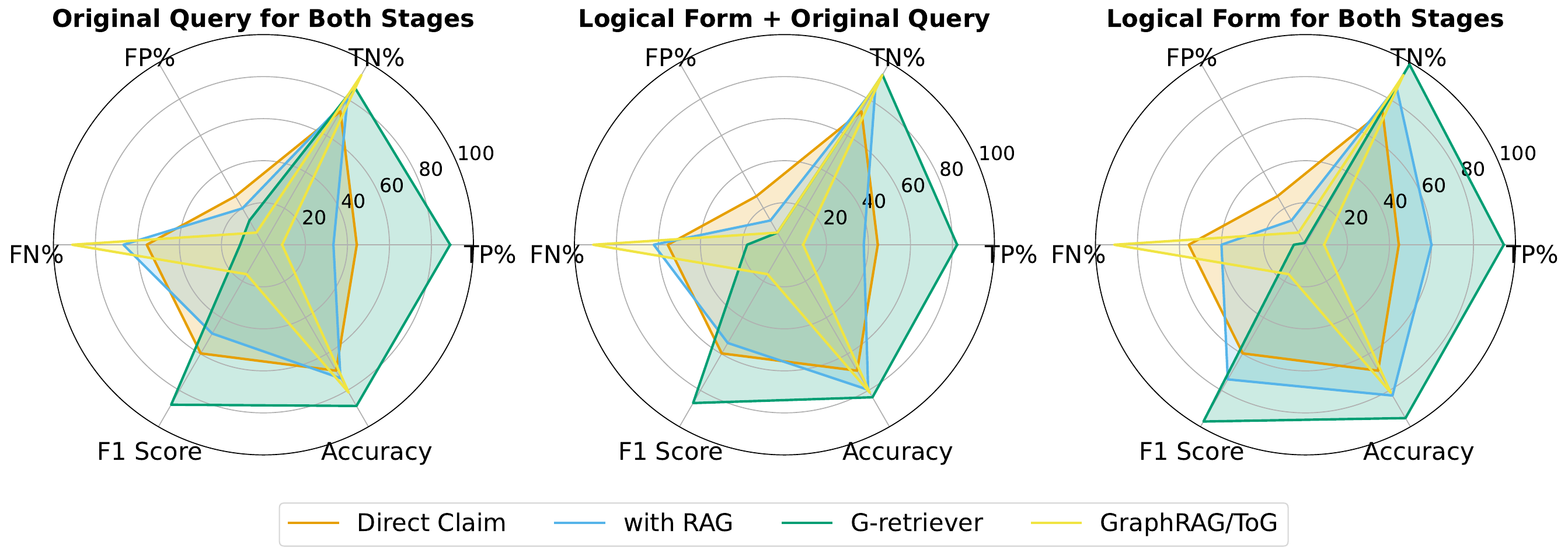}
  }
  \caption{KG-FPQ dataset: comparison of performance metrics across different retrieval methods using logical forms and/or original queries.}
  \label{fig:radar}
\end{figure*}
\begin{table*}[h!]
\small
\centering
\scalebox{0.8}{
\begin{tabular}{lcccccc}
\hline
\textbf{Method} & \textbf{Accuracy} & \textbf{Number of tokens} & \textbf{Running time*} & \textbf{Model agnostic} & \textbf{Black-box Compatible} \\
\hline
Contrastive& 84.8 & Original Query + Reasoning Step & Context Retrieval Time & Agnostic to & No \\
Decoding && (Length $\gg$ Logical Form) & + 10.6s & White Box Models & \\
Our Method & 89.5 & Original Query + Logical Form & Context Retrieval Time& Yes & Yes \\
&&&+ 0.6s && \\
\hline
\end{tabular}
}
\caption{Comparison of performance and efficiency between contrastive decoding and our method on the KG-FPQ dataset. *Average running time of each query on NVIDIA RTX A6000 GPU using Llama-3.1-8B Instruct model. Both methods require context retrieval.}
\label{tab:efficiency_comparison}
\end{table*}

\subsubsection{Hallucination Mitigation Methods}\label{sec:4_3}
Having used logical forms to improve query structuring and false premise detection, we wish to illustrate how our logical form-based method further reduces hallucinations. We consider the following methods as our hallucination mitigation baselines, which are all inference-time hallucination mitigation strategies that do not require access to logits or internal model weights that operate exclusively at the input level, ensuring a fair comparison:

1) \textbf{DirectAsk}:  Directly query the LLMs for an answer without additional processing or external retrieval. This approach relies on the model’s internal knowledge and reasoning capabilities to handle potential false premises.

2) \textbf{Prompt}: We encourage the LLM to assess potential false premises before generating a response by appending the following prompt to the original query:
\textit{This question may contain a fasle premise. [query]}

3) \textbf{Majority Vote (MajVote)}: 
We prompt the LLM three times with the same prompt and select the most frequent response as the final answer. This method improves reliability by reducing the impact of any single erroneous or hallucinated response.
from LLM.

4) \textbf{Perplexity AI}\footnote{https://www.perplexity.ai}:
Utilizes a search engine to retrieve and incorporate real-time information from the web, enabling it to provide answers based on the latest available web data. We use the version powered by GPT-4-Omni.

5) \textbf{Direct RAG}:
Retrieves relevant entities from the knowledge graph and provides them as context to the LLM alongside the original query. This approach augments the model's internal knowledge directly with external information to improve answer accuracy and grounding.
 
We report the performances of the following LLMs: GPT-4o-mini \citep{openai2024gpt4technicalreport}, GPT-3.5-turbo \citep{OpenAI_GPT35_2023}, Llama-3.1-8B-Instruct \citep{grattafiori2024llama3herdmodels}, Mistral-7B-Instruct-v0.2 \citep{jiang2023mistral7b}, Qwen2.5-7B-Instruct \citep{qwen2025qwen25technicalreport}, and Qwen-1.5-7b-chat \citep{bai2023qwentechnicalreport}. \update{To better understand where the performance gains come from, we provide a per-model breakdown of error corrections and premise-specific changes (false- vs. true-premise queries) in Appendix~\S \ref{app:premise_breakdown}}.

\begin{table}[h]
\centering
\begin{tabular}{l|ccccc}
\hline
\textbf{Models} & \textbf{DirectAsk} & \textbf{Prompt} & \textbf{MajVote} & \textbf{DirectRAG} & \textbf{Ours} \\
\hline
GPT-4o-mini & 83.8 & 92.4 & 86.7 & 90.5 & \textbf{92.4} \\
GPT-3.5 & 93.3 & 93.3 & 92.4 & 90.5 & \textbf{94.3} \\
LLama-3.1 & 86.7 & 86.7 & 89.5 & 88.6 & \textbf{89.5} \\
Mistral-7B & 87.6 & 86.7 & 87.6 & 71.4 & \textbf{89.5} \\
Qwen2.5 & 92.4 & 86.7 & 92.4 & 92.4 & \textbf{95.2} \\
Qwen1.5 & 89.5 & 90.5 & 90.5 & 82.9 & \textbf{91.4} \\
\hline
\hline
Perplexity AI & \multicolumn{5}{c}{91.4} \\
\hline
\end{tabular}
\caption{Comparison of accuracy (\%) of different hallucination mitigation methods.}
\label{tab:hallucination_mitigation}
\end{table}
 \textbf{Metrics}.
 We evaluate question-answering accuracy on the hallucination mitigation task.
Accuracy is calculated by string matching the responses of LLMs: for TPQs, answering “Yes” is considered correct; for FPQs, answering “No” is considered correct.
 
\section{Discussion}
\update{We show the result of the false premise detection task in Tab.\ref{tab:false_premise_detection_perf}, \ref{tab:creak}, and \ref{tab:fever} for the KG-FPQ, CREAK and FEVER dataset, respectively}. The SAC3 baseline result is shown in Appendix \S~\ref{sec:sac3}. Tab.~\ref{tab:hallucination_mitigation} presents the hallucination mitigation result.

\textbf{Using logical forms helps better identify false premises in the questions}. As shown in Tab.~\ref{tab:false_premise_detection_perf}, for all three retrievers, explicitly incorporating logical forms into both retrieval and false premise detection stages significantly improves the identification of false premises. Sole reliance on original queries, even though potentially yielding high accuracy, tends to neglect accurate false premise identification, underscoring the importance of utilizing structured logical forms for tasks prioritizing precise false premise detection.

For the KG-FPQ dataset, among different types of retrievers, when using logical forms in both the retrieval and false premise detection stages, the G-retriever method achieves the highest TPR at 94.44\%, demonstrating a strong capability in accurately identifying questions containing false premises. Notably, this method also achieves the highest F1 score (97.12\%), indicating an optimal balance between precision and recall. Although the ToG method exhibits the highest TNR of 93.33\%, it significantly underperforms in TPR and overall F1 score (16.16\%), suggesting limited effectiveness in correctly identifying false premises.

Notably, when original queries are used in either retrieval, false premise detection, or both stages,
despite achieving reasonable accuracy (73.33\% and 79.69\%), with RAG method shows significantly lower TPR (33.33\% and 37.78\%) compared to the first configuration. This suggests that relying on original queries alone, or in combination with logical forms in only one stage for detection, can achieve high accuracy due to correctly identifying negatives, it is less effective at capturing false premises, which is the primary focus of our task.

Similarly, for the CREAK dataset, according to Tab.~\ref{tab:creak},
using logical forms in both retrieval and detection stages consistently boosts performance versus operating on the original query. The gains are most pronounced for G-retriever: F1 score increases from 71.1\% to 90.4\% and accuracy from 74.5\% to 90.5\%, driven by a large increase in TPR (62.3\% to 88.4\%). GraphRAG/ToG already performs strongly with original queries, but still benefits from using logical forms, reaching the best overall scores (F1 92.1\% and Acc 92.0\%) with higher TPR (89.9\% to 92.8\%) while keeping FPR modest. The mixed configuration for retrieval mainly helps G-retriever
but is less reliable than using logical forms in both stages, underscoring that structured queries plus explicit false premise detection are jointly necessary. Logical forms improves evidence targeting and makes contradictions salient to LLMs. Overall, the results indicate that logic-aware framework is crucial for converting strong retrieval into consistent end-to-end factuality.

\update{According to Tab.~\ref{tab:fever}, the proposed design also yields consistent performance gains on natural language knowledge base, indicating that incorporating logical forms in both the premise detection and LLM response stages remains effective when the knowledge source is less structured. This suggests that the benefits of logical-form guided reasoning stem from improved premise understanding and response control, rather than reliance on a specific knowledge representation.}

\textbf{Explicitly detecting and informing LLMs false premise mitigates hallucination}, as demonstrated in Tab.~\ref{tab:hallucination_mitigation}. Our proposed method, which directly communicates the presence of false premises to the models, achieves the highest accuracy: 92.4\% with GPT-4o-mini, 94.3\% with GPT-3.5, 95.2\% with Qwen2.5, and 91.4\% with Qwen-1.5. This performance surpasses alternative approaches such as \textit{Direct Ask}, \textit{Prompt}, \textit{Majority Vote}, \textit{DirectRAG}, and \textit{Perplexity AI}. 

\textit{Majority Vote} does not perform well, likely due to hallucination snowballing, where repeated querying amplifies errors rather than correcting them. Additionally, while the \textit{Prompt} method warns the model about potential false premises, it does not specifically tell the LLM which one contains false premises, negatively impacts performance on questions with valid premises, causes unnecessary cautiousness and reduces the model’s ability to provide direct and confident answers. Besides, \textit{Perplexity AI} does not perform as well potentially because the query format does not align well with graph data, leading to suboptimal retrieval of relevant information for certain types of questions. These findings emphasize the importance of tailoring hallucination mitigation strategies to both the model's reasoning process and the nature of the queries.

\textit{Direct RAG} retrieves and feeds raw evidence to LLMs without explicitly structuring the reasoning process or highlighting inconsistencies between the query and retrieved facts. As a result, the model may surface relevant but semantically unaligned information, leading to shallow retrieval-based responses rather than true logical verification. In contrast, logical-form RAG + explicit false premise signaling forms a structured reasoning process: it decomposes the claim into logical predicates and explicitly indicates when a premise conflicts with retrieved evidence. This guides LLMs to perform fact-level reasoning and contradiction handling, reducing overreliance on surface overlap and improving factual precision and interpretability.

\update{Beyond hallucination mitigation, the proposed premise verification mechanism can be extended to sensitive or controversial topics, where unverified premises may amplify misinformation or harmful narratives. By explicitly detecting unsupported assumptions prior to response generation, the method offers a principled way to prevent models from uncritically engaging with inaccurate or inflammatory premises, enabling safer and more grounded interactions in high-risk domains. This suggests a broader role for retrieval-augmented logical reasoning as a lightweight safeguard for responsible deployment, especially in scenarios where factual grounding is essential before engaging in downstream reasoning or dialogue.}

\begin{figure}[h]
    \small
    \centering
    \scalebox{0.4}{
\includegraphics[width=\linewidth]{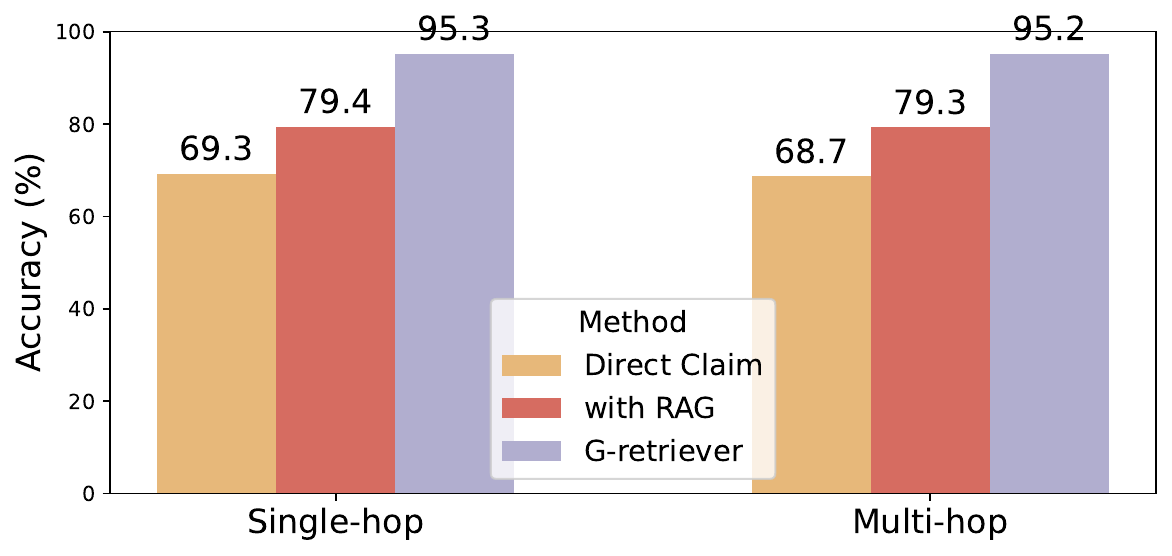}
}
    \caption{GPT-4o-mini and G-retriver: False premise detection accuracy across single-hop and multi-hop queries on the KG-FPQ dataset. Using logical form-based RAG mainly helps detect false premises in multi-hop questions.}
    \label{fig:num_hop}
\end{figure}
\textbf{Our approach mostly improves false premise detection performance on multi-hop questions}, according to
Fig.~\ref{fig:num_hop}.
The incorporation of logical form-based RAG leads to notable performance gains compared to direct claim evaluation. Specifically, while single-hop questions see moderate improvement, multi-hop questions benefit more, with false premise detection performance increasing from 68.7\% in the direct claim setting to 79.3\% with RAG and further to 95.2\% when using the G-retriever. These results suggest that leveraging retrieval mechanisms enhances reasoning over multiple pieces of evidence, reinforcing the importance of retrieval-augmented methods for complex question-answering tasks.
We present a case study to illustrate how our method improves performance on multi-hop questions in 
Appendix \S~\ref{sec:case_study}.

\subsection{Computational Cost Analysis}
In Tab.~\ref{tab:efficiency_comparison}, we compare our method with the post-hoc Contrastive Decoding \citep{shi2023trustingevidencehallucinatecontextaware} approach in terms of computational efficiency and model compatibility (accuracy result based on Llama-3.1-8B).
Our method reduces running time, uses fewer tokens by leveraging logical forms, and supports both model-agnostic and black-box settings.
The proposed method introduces only a modest computational overhead, adding $\mathcal{O}(k + s)$ complexity to the retriever, where $k$ is the number of retrieved candidates and $s$ is the feature size used for logical-form reasoning and premise detection. In practice, this addition remains lightweight since most of the cost lies in the embedding model inference, which requires approximately 0.335 TFLOPs per example, while the remaining steps—retrieval, logical form conversion, and false-premise detection—are implemented as efficient API calls. 
In contrast, post-hoc methods rely on fine-tuning and lack general applicability across different model architectures.
We also include performance comparison of Contrastive Decoding with other LLMs in Appendix \S~\ref{sec:add_posthoc}.
\subsubsection{Significance Test}\label{sec:ttest}
To ensure that the reported performance differences are statistically meaningful, we conduct paired t-tests across model configurations. The results, summarized in Table \ref{tab:retriever_pvalues} and Table \ref{tab:retriever_config_pvalues} , confirm that the observed improvements are statistically significant (p < 0.05) in most comparisons.
Table \ref{tab:retriever_pvalues} evaluates overall differences among performances of Direct Claim, with RAG, and G-Retriever on the KG-FPQ dataset.
Table \ref{tab:retriever_config_pvalues} focuses on intra-method variations, contrasting setups that use logical forms in both stages (GG), only one stage (GO), or none (OO). The results confirm that applying logical forms consistently across both retrieval and detection stages yields significantly higher performance, reinforcing the benefit of structured reasoning input.

\begin{table}[h]
\centering
\small

\begin{minipage}{0.48\textwidth}
\centering
\begin{tabular}{lc}
\hline
\textbf{Comparison} & \textbf{p-value} \\
\hline
G-Retriever -- wRAG & 0.04 \\
Direct Claim -- G-Retriever & 0.001 \\
wRAG -- G-Retriever & 0.02 \\
\hline
\end{tabular}
\caption{Statistical significance comparison across retrieval methods.}
\label{tab:retriever_pvalues}
\end{minipage}
\hfill
\begin{minipage}{0.48\textwidth}
\centering
\begin{tabular}{lc}
\hline
\textbf{Comparison} & \textbf{p-value} \\
\hline
G-Retriever (GG) -- G-Retriever (GO) & $<0.001$ \\
G-Retriever (GG) -- G-Retriever (OO) & $<0.001$ \\
wRAG (GG) -- wRAG (GO) & 0.01 \\
\hline
\end{tabular}
\caption{Significance comparison of configurations within retrieval methods.}
\label{tab:retriever_config_pvalues}
\end{minipage}

\end{table}

\section{Conclusion}
We propose a retrieval-augmented logical reasoning framework that detects false premises to mitigate LLM hallucinations. Our method explicitly detects and signals false premises, overcoming key limitations of current approaches that rely on model parameters or post-hoc corrections. By incorporating upfront false premise detection, we prevent hallucinations without requiring output generation or model logit access. Results show logical forms significantly improve false premise identification, especially for multi-hop reasoning questions. Our approach enhances LLM robustness by providing a structured mechanism to detect and handle misleading inputs before they influence downstream responses. This reinforces the importance of structured reasoning techniques in improving model reliability.

\bibliography{main}

@misc{openai2024gpt4technicalreport,
      title={GPT-4 Technical Report}, 
      author={OpenAI},
      year={2024},
      eprint={2303.08774},
      archivePrefix={arXiv},
      primaryClass={cs.CL},
      url={https://arxiv.org/abs/2303.08774}, 
}

@misc{zhu2024kgfpqevaluatingfactualityhallucination,
      title={KG-FPQ: Evaluating Factuality Hallucination in LLMs with Knowledge Graph-based False Premise Questions}, 
      author={Yanxu Zhu and Jinlin Xiao and Yuhang Wang and Jitao Sang},
      year={2024},
      eprint={2407.05868},
      archivePrefix={arXiv},
      primaryClass={cs.CL},
      url={https://arxiv.org/abs/2407.05868}, 
}

@misc{liu2024knowledgegraphenhancedlargelanguage,
      title={Knowledge Graph-Enhanced Large Language Models via Path Selection}, 
      author={Haochen Liu and Song Wang and Yaochen Zhu and Yushun Dong and Jundong Li},
      year={2024},
      eprint={2406.13862},
      archivePrefix={arXiv},
      primaryClass={cs.CL},
      url={https://arxiv.org/abs/2406.13862}, 
}

@inproceedings{zheng-etal-2024-evidence,
    title = "Evidence Retrieval is almost All You Need for Fact Verification",
    author = "Zheng, Liwen  and
      Li, Chaozhuo  and
      Zhang, Xi  and
      Shang, Yu-Ming  and
      Huang, Feiran  and
      Jia, Haoran",
    editor = "Ku, Lun-Wei  and
      Martins, Andre  and
      Srikumar, Vivek",
    booktitle = "Findings of the Association for Computational Linguistics: ACL 2024",
    month = aug,
    year = "2024",
    address = "Bangkok, Thailand",
    publisher = "Association for Computational Linguistics",
    url = "https://aclanthology.org/2024.findings-acl.551/",
    doi = "10.18653/v1/2024.findings-acl.551",
    pages = "9274--9281"
}

@misc{pan2023qacheckdemonstrationquestionguidedmultihop,
      title={QACHECK: A Demonstration System for Question-Guided Multi-Hop Fact-Checking}, 
      author={Liangming Pan and Xinyuan Lu and Min-Yen Kan and Preslav Nakov},
      year={2023},
      eprint={2310.07609},
      archivePrefix={arXiv},
      primaryClass={cs.CL},
      url={https://arxiv.org/abs/2310.07609}, 
}

@misc{yu2022crepeopendomainquestionanswering,
      title={CREPE: Open-Domain Question Answering with False Presuppositions}, 
      author={Xinyan Velocity Yu and Sewon Min and Luke Zettlemoyer and Hannaneh Hajishirzi},
      year={2022},
      eprint={2211.17257},
      archivePrefix={arXiv},
      primaryClass={cs.CL},
      url={https://arxiv.org/abs/2211.17257}, 
}

@misc{vu2023freshllmsrefreshinglargelanguage,
      title={FreshLLMs: Refreshing Large Language Models with Search Engine Augmentation}, 
      author={Tu Vu and Mohit Iyyer and Xuezhi Wang and Noah Constant and Jerry Wei and Jason Wei and Chris Tar and Yun-Hsuan Sung and Denny Zhou and Quoc Le and Thang Luong},
      year={2023},
      eprint={2310.03214},
      archivePrefix={arXiv},
      primaryClass={cs.CL},
      url={https://arxiv.org/abs/2310.03214}, 
}

@misc{yuan2024whispersshakefoundationsanalyzing,
      title={Whispers that Shake Foundations: Analyzing and Mitigating False Premise Hallucinations in Large Language Models}, 
      author={Hongbang Yuan and Pengfei Cao and Zhuoran Jin and Yubo Chen and Daojian Zeng and Kang Liu and Jun Zhao},
      year={2024},
      eprint={2402.19103},
      archivePrefix={arXiv},
      primaryClass={cs.CL},
      url={https://arxiv.org/abs/2402.19103}, 
}

@misc{varshney2023stitchtimesavesnine,
      title={A Stitch in Time Saves Nine: Detecting and Mitigating Hallucinations of LLMs by Validating Low-Confidence Generation}, 
      author={Neeraj Varshney and Wenlin Yao and Hongming Zhang and Jianshu Chen and Dong Yu},
      year={2023},
      eprint={2307.03987},
      archivePrefix={arXiv},
      primaryClass={cs.CL},
      url={https://arxiv.org/abs/2307.03987}, 
}

@misc{zheng2023judgingllmasajudgemtbenchchatbot,
      title={Judging LLM-as-a-Judge with MT-Bench and Chatbot Arena}, 
      author={Lianmin Zheng and Wei-Lin Chiang and Ying Sheng and Siyuan Zhuang and Zhanghao Wu and Yonghao Zhuang and Zi Lin and Zhuohan Li and Dacheng Li and Eric P. Xing and Hao Zhang and Joseph E. Gonzalez and Ion Stoica},
      year={2023},
      eprint={2306.05685},
      archivePrefix={arXiv},
      primaryClass={cs.CL},
      url={https://arxiv.org/abs/2306.05685}, 
}

@misc{manakul2023selfcheckgptzeroresourceblackboxhallucination,
      title={SelfCheckGPT: Zero-Resource Black-Box Hallucination Detection for Generative Large Language Models}, 
      author={Potsawee Manakul and Adian Liusie and Mark J. F. Gales},
      year={2023},
      eprint={2303.08896},
      archivePrefix={arXiv},
      primaryClass={cs.CL},
      url={https://arxiv.org/abs/2303.08896}, 
}

@inproceedings{10.1145/3637528.3671796,
author = {Snyder, Ben and Moisescu, Marius and Zafar, Muhammad Bilal},
title = {On Early Detection of Hallucinations in Factual Question Answering},
year = {2024},
isbn = {9798400704901},
publisher = {Association for Computing Machinery},
address = {New York, NY, USA},
url = {https://doi.org/10.1145/3637528.3671796},
doi = {10.1145/3637528.3671796},
booktitle = {Proceedings of the 30th ACM SIGKDD Conference on Knowledge Discovery and Data Mining},
pages = {2721–2732},
numpages = {12},
location = {Barcelona, Spain},
series = {KDD '24}
}

@article{Huang_2025,
   title={A Survey on Hallucination in Large Language Models: Principles, Taxonomy, Challenges, and Open Questions},
   volume={43},
   ISSN={1558-2868},
   url={http://dx.doi.org/10.1145/3703155},
   DOI={10.1145/3703155},
   number={2},
   journal={ACM Transactions on Information Systems},
   publisher={Association for Computing Machinery (ACM)},
   author={Huang, Lei and Yu, Weijiang and Ma, Weitao and Zhong, Weihong and Feng, Zhangyin and Wang, Haotian and Chen, Qianglong and Peng, Weihua and Feng, Xiaocheng and Qin, Bing and Liu, Ting},
   year={2025},
   month=jan, pages={1–55} }

@inproceedings{snyder2024early,
  title={On early detection of hallucinations in factual question answering},
  author={Snyder, Ben and Moisescu, Marius and Zafar, Muhammad Bilal},
  booktitle={Proceedings of the 30th ACM SIGKDD Conference on Knowledge Discovery and Data Mining},
  pages={2721--2732},
  year={2024}
}

@misc{pal2023medhaltmedicaldomainhallucination,
      title={Med-HALT: Medical Domain Hallucination Test for Large Language Models}, 
      author={Ankit Pal and Logesh Kumar Umapathi and Malaikannan Sankarasubbu},
      year={2023},
      eprint={2307.15343},
      archivePrefix={arXiv},
      primaryClass={cs.CL},
      url={https://arxiv.org/abs/2307.15343}, 
}

@misc{hu2023wontfooledagainanswering,
      title={Won't Get Fooled Again: Answering Questions with False Premises}, 
      author={Shengding Hu and Yifan Luo and Huadong Wang and Xingyi Cheng and Zhiyuan Liu and Maosong Sun},
      year={2023},
      eprint={2307.02394},
      archivePrefix={arXiv},
      primaryClass={cs.CL},
      url={https://arxiv.org/abs/2307.02394}, 
}

@misc{pezeshkpour2023measuringmodifyingfactualknowledge,
      title={Measuring and Modifying Factual Knowledge in Large Language Models}, 
      author={Pouya Pezeshkpour},
      year={2023},
      eprint={2306.06264},
      archivePrefix={arXiv},
      primaryClass={cs.CL},
      url={https://arxiv.org/abs/2306.06264}, 
}

@misc{liu2024selfcontradictoryreasoningevaluationdetection,
      title={Self-Contradictory Reasoning Evaluation and Detection}, 
      author={Ziyi Liu and Soumya Sanyal and Isabelle Lee and Yongkang Du and Rahul Gupta and Yang Liu and Jieyu Zhao},
      year={2024},
      eprint={2311.09603},
      archivePrefix={arXiv},
      primaryClass={cs.CL},
      url={https://arxiv.org/abs/2311.09603}, 
}

@misc{shi2023trustingevidencehallucinatecontextaware,
      title={Trusting Your Evidence: Hallucinate Less with Context-aware Decoding}, 
      author={Weijia Shi and Xiaochuang Han and Mike Lewis and Yulia Tsvetkov and Luke Zettlemoyer and Scott Wen-tau Yih},
      year={2023},
      eprint={2305.14739},
      archivePrefix={arXiv},
      primaryClass={cs.CL},
      url={https://arxiv.org/abs/2305.14739}, 
}

@misc{chuang2024doladecodingcontrastinglayers,
      title={DoLa: Decoding by Contrasting Layers Improves Factuality in Large Language Models}, 
      author={Yung-Sung Chuang and Yujia Xie and Hongyin Luo and Yoon Kim and James Glass and Pengcheng He},
      year={2024},
      eprint={2309.03883},
      archivePrefix={arXiv},
      primaryClass={cs.CL},
      url={https://arxiv.org/abs/2309.03883}, 
}

@misc{sun2024thinkongraphdeepresponsiblereasoning,
      title={Think-on-Graph: Deep and Responsible Reasoning of Large Language Model on Knowledge Graph}, 
      author={Jiashuo Sun and Chengjin Xu and Lumingyuan Tang and Saizhuo Wang and Chen Lin and Yeyun Gong and Lionel M. Ni and Heung-Yeung Shum and Jian Guo},
      year={2024},
      eprint={2307.07697},
      archivePrefix={arXiv},
      primaryClass={cs.CL},
      url={https://arxiv.org/abs/2307.07697}, 
}

@misc{he2024gretrieverretrievalaugmentedgenerationtextual,
      title={G-Retriever: Retrieval-Augmented Generation for Textual Graph Understanding and Question Answering}, 
      author={Xiaoxin He and Yijun Tian and Yifei Sun and Nitesh V. Chawla and Thomas Laurent and Yann LeCun and Xavier Bresson and Bryan Hooi},
      year={2024},
      eprint={2402.07630},
      archivePrefix={arXiv},
      primaryClass={cs.LG},
      url={https://arxiv.org/abs/2402.07630}, 
}

@misc{mavromatis2024gnnraggraphneuralretrieval,
      title={GNN-RAG: Graph Neural Retrieval for Large Language Model Reasoning}, 
      author={Costas Mavromatis and George Karypis},
      year={2024},
      eprint={2405.20139},
      archivePrefix={arXiv},
      primaryClass={cs.CL},
      url={https://arxiv.org/abs/2405.20139}, 
}

@misc{zhang2023languagemodelhallucinationssnowball,
      title={How Language Model Hallucinations Can Snowball}, 
      author={Muru Zhang and Ofir Press and William Merrill and Alisa Liu and Noah A. Smith},
      year={2023},
      eprint={2305.13534},
      archivePrefix={arXiv},
      primaryClass={cs.CL},
      url={https://arxiv.org/abs/2305.13534}, 
}

@misc{touvron2023llama2openfoundation,
      title={Llama 2: Open Foundation and Fine-Tuned Chat Models}, 
      author={Hugo Touvron and Louis Martin and Kevin Stone and Peter Albert and Amjad Almahairi and Yasmine Babaei and Nikolay Bashlykov and Soumya Batra and Prajjwal Bhargava and Shruti Bhosale and Dan Bikel and Lukas Blecher and Cristian Canton Ferrer and Moya Chen and Guillem Cucurull and David Esiobu and Jude Fernandes and Jeremy Fu and Wenyin Fu and Brian Fuller and Cynthia Gao and Vedanuj Goswami and Naman Goyal and Anthony Hartshorn and Saghar Hosseini and Rui Hou and Hakan Inan and Marcin Kardas and Viktor Kerkez and Madian Khabsa and Isabel Kloumann and Artem Korenev and Punit Singh Koura and Marie-Anne Lachaux and Thibaut Lavril and Jenya Lee and Diana Liskovich and Yinghai Lu and Yuning Mao and Xavier Martinet and Todor Mihaylov and Pushkar Mishra and Igor Molybog and Yixin Nie and Andrew Poulton and Jeremy Reizenstein and Rashi Rungta and Kalyan Saladi and Alan Schelten and Ruan Silva and Eric Michael Smith and Ranjan Subramanian and Xiaoqing Ellen Tan and Binh Tang and Ross Taylor and Adina Williams and Jian Xiang Kuan and Puxin Xu and Zheng Yan and Iliyan Zarov and Yuchen Zhang and Angela Fan and Melanie Kambadur and Sharan Narang and Aurelien Rodriguez and Robert Stojnic and Sergey Edunov and Thomas Scialom},
      year={2023},
      eprint={2307.09288},
      archivePrefix={arXiv},
      primaryClass={cs.CL},
      url={https://arxiv.org/abs/2307.09288}, 
}

@misc{lee2023factualityenhancedlanguagemodels,
      title={Factuality Enhanced Language Models for Open-Ended Text Generation}, 
      author={Nayeon Lee and Wei Ping and Peng Xu and Mostofa Patwary and Pascale Fung and Mohammad Shoeybi and Bryan Catanzaro},
      year={2023},
      eprint={2206.04624},
      archivePrefix={arXiv},
      primaryClass={cs.CL},
      url={https://arxiv.org/abs/2206.04624}, 
}

@misc{chen2024alpagasustrainingbetteralpaca,
      title={AlpaGasus: Training A Better Alpaca with Fewer Data}, 
      author={Lichang Chen and Shiyang Li and Jun Yan and Hai Wang and Kalpa Gunaratna and Vikas Yadav and Zheng Tang and Vijay Srinivasan and Tianyi Zhou and Heng Huang and Hongxia Jin},
      year={2024},
      eprint={2307.08701},
      archivePrefix={arXiv},
      primaryClass={cs.CL},
      url={https://arxiv.org/abs/2307.08701}, 
}

@misc{cao2024instructionmininginstructiondata,
      title={Instruction Mining: Instruction Data Selection for Tuning Large Language Models}, 
      author={Yihan Cao and Yanbin Kang and Chi Wang and Lichao Sun},
      year={2024},
      eprint={2307.06290},
      archivePrefix={arXiv},
      primaryClass={cs.CL},
      url={https://arxiv.org/abs/2307.06290}, 
}

@misc{zhang2023sirenssongaiocean,
      title={Siren's Song in the AI Ocean: A Survey on Hallucination in Large Language Models}, 
      author={Yue Zhang and Yafu Li and Leyang Cui and Deng Cai and Lemao Liu and Tingchen Fu and Xinting Huang and Enbo Zhao and Yu Zhang and Yulong Chen and Longyue Wang and Anh Tuan Luu and Wei Bi and Freda Shi and Shuming Shi},
      year={2023},
      eprint={2309.01219},
      archivePrefix={arXiv},
      primaryClass={cs.CL},
      url={https://arxiv.org/abs/2309.01219}, 
}

@misc{radhakrishnan2023questiondecompositionimprovesfaithfulness,
      title={Question Decomposition Improves the Faithfulness of Model-Generated Reasoning}, 
      author={Ansh Radhakrishnan and Karina Nguyen and Anna Chen and Carol Chen and Carson Denison and Danny Hernandez and Esin Durmus and Evan Hubinger and Jackson Kernion and Kamilė Lukošiūtė and Newton Cheng and Nicholas Joseph and Nicholas Schiefer and Oliver Rausch and Sam McCandlish and Sheer El Showk and Tamera Lanham and Tim Maxwell and Venkatesa Chandrasekaran and Zac Hatfield-Dodds and Jared Kaplan and Jan Brauner and Samuel R. Bowman and Ethan Perez},
      year={2023},
      eprint={2307.11768},
      archivePrefix={arXiv},
      primaryClass={cs.CL},
      url={https://arxiv.org/abs/2307.11768}, 
}

@misc{wei2024simplesyntheticdatareduces,
      title={Simple synthetic data reduces sycophancy in large language models}, 
      author={Jerry Wei and Da Huang and Yifeng Lu and Denny Zhou and Quoc V. Le},
      year={2024},
      eprint={2308.03958},
      archivePrefix={arXiv},
      primaryClass={cs.CL},
      url={https://arxiv.org/abs/2308.03958}, 
}

@misc{dhuliawala2023chainofverificationreduceshallucinationlarge,
      title={Chain-of-Verification Reduces Hallucination in Large Language Models}, 
      author={Shehzaad Dhuliawala and Mojtaba Komeili and Jing Xu and Roberta Raileanu and Xian Li and Asli Celikyilmaz and Jason Weston},
      year={2023},
      eprint={2309.11495},
      archivePrefix={arXiv},
      primaryClass={cs.CL},
      url={https://arxiv.org/abs/2309.11495}, 
}

@misc{xu2025decopromptdecodingprompts,
      title={DecoPrompt : Decoding Prompts Reduces Hallucinations when Large Language Models Meet False Premises}, 
      author={Nan Xu and Xuezhe Ma},
      year={2025},
      eprint={2411.07457},
      archivePrefix={arXiv},
      primaryClass={cs.CL},
      url={https://arxiv.org/abs/2411.07457}, 
}

@misc{brown2020languagemodelsfewshotlearners,
      title={Language Models are Few-Shot Learners}, 
      author={Tom B. Brown and Benjamin Mann and Nick Ryder and Melanie Subbiah and Jared Kaplan and Prafulla Dhariwal and Arvind Neelakantan and Pranav Shyam and Girish Sastry and Amanda Askell and Sandhini Agarwal and Ariel Herbert-Voss and Gretchen Krueger and Tom Henighan and Rewon Child and Aditya Ramesh and Daniel M. Ziegler and Jeffrey Wu and Clemens Winter and Christopher Hesse and Mark Chen and Eric Sigler and Mateusz Litwin and Scott Gray and Benjamin Chess and Jack Clark and Christopher Berner and Sam McCandlish and Alec Radford and Ilya Sutskever and Dario Amodei},
      year={2020},
      eprint={2005.14165},
      archivePrefix={arXiv},
      primaryClass={cs.CL},
      url={https://arxiv.org/abs/2005.14165}, 
}

@misc{wei2023chainofthoughtpromptingelicitsreasoning,
      title={Chain-of-Thought Prompting Elicits Reasoning in Large Language Models}, 
      author={Jason Wei and Xuezhi Wang and Dale Schuurmans and Maarten Bosma and Brian Ichter and Fei Xia and Ed Chi and Quoc Le and Denny Zhou},
      year={2023},
      eprint={2201.11903},
      archivePrefix={arXiv},
      primaryClass={cs.CL},
      url={https://arxiv.org/abs/2201.11903}, 
}

@misc{bai2023qwentechnicalreport,
      title={Qwen Technical Report}, 
      author={Jinze Bai and Shuai Bai and Yunfei Chu and Zeyu Cui and Kai Dang and Xiaodong Deng and Yang Fan and Wenbin Ge and Yu Han and Fei Huang and Binyuan Hui and Luo Ji and Mei Li and Junyang Lin and Runji Lin and Dayiheng Liu and Gao Liu and Chengqiang Lu and Keming Lu and Jianxin Ma and Rui Men and Xingzhang Ren and Xuancheng Ren and Chuanqi Tan and Sinan Tan and Jianhong Tu and Peng Wang and Shijie Wang and Wei Wang and Shengguang Wu and Benfeng Xu and Jin Xu and An Yang and Hao Yang and Jian Yang and Shusheng Yang and Yang Yao and Bowen Yu and Hongyi Yuan and Zheng Yuan and Jianwei Zhang and Xingxuan Zhang and Yichang Zhang and Zhenru Zhang and Chang Zhou and Jingren Zhou and Xiaohuan Zhou and Tianhang Zhu},
      year={2023},
      eprint={2309.16609},
      archivePrefix={arXiv},
      primaryClass={cs.CL},
      url={https://arxiv.org/abs/2309.16609}, 
}

@misc{OpenAI_GPT35_2023,
  author = {OpenAI},
  title = {GPT-3.5-Turbo: Large Language Model},
  year = {2023},
  url = {https://platform.openai.com/docs/models/gpt-3-5},
  note = {Accessed: 2024-03-18}
}

@misc{edge2025localglobalgraphrag,
      title={From Local to Global: A Graph RAG Approach to Query-Focused Summarization}, 
      author={Darren Edge and Ha Trinh and Newman Cheng and Joshua Bradley and Alex Chao and Apurva Mody and Steven Truitt and Dasha Metropolitansky and Robert Osazuwa Ness and Jonathan Larson},
      year={2025},
      eprint={2404.16130},
      archivePrefix={arXiv},
      primaryClass={cs.CL},
      url={https://arxiv.org/abs/2404.16130}, 
}

@misc{jiang2023mistral7b,
      title={Mistral 7B}, 
      author={Albert Q. Jiang and Alexandre Sablayrolles and Arthur Mensch and Chris Bamford and Devendra Singh Chaplot and Diego de las Casas and Florian Bressand and Gianna Lengyel and Guillaume Lample and Lucile Saulnier and Lélio Renard Lavaud and Marie-Anne Lachaux and Pierre Stock and Teven Le Scao and Thibaut Lavril and Thomas Wang and Timothée Lacroix and William El Sayed},
      year={2023},
      eprint={2310.06825},
      archivePrefix={arXiv},
      primaryClass={cs.CL},
      url={https://arxiv.org/abs/2310.06825}, 
}

@misc{liu2024dellmadecisionmakinguncertainty,
      title={DeLLMa: Decision Making Under Uncertainty with Large Language Models}, 
      author={Ollie Liu and Deqing Fu and Dani Yogatama and Willie Neiswanger},
      year={2024},
      eprint={2402.02392},
      archivePrefix={arXiv},
      primaryClass={cs.AI},
      url={https://arxiv.org/abs/2402.02392}, 
}

@misc{kim2021linguistinventedlightbulbpresupposition,
      title={Which Linguist Invented the Lightbulb? Presupposition Verification for Question-Answering}, 
      author={Najoung Kim and Ellie Pavlick and Burcu Karagol Ayan and Deepak Ramachandran},
      year={2021},
      eprint={2101.00391},
      archivePrefix={arXiv},
      primaryClass={cs.CL},
      url={https://arxiv.org/abs/2101.00391}, 
}

@inproceedings{Olausson_2023,
   title={LINC: A Neurosymbolic Approach for Logical Reasoning by Combining Language Models with First-Order Logic Provers},
   url={http://dx.doi.org/10.18653/v1/2023.emnlp-main.313},
   DOI={10.18653/v1/2023.emnlp-main.313},
   booktitle={Proceedings of the 2023 Conference on Empirical Methods in Natural Language Processing},
   publisher={Association for Computational Linguistics},
   author={Olausson, Theo and Gu, Alex and Lipkin, Ben and Zhang, Cedegao and Solar-Lezama, Armando and Tenenbaum, Joshua and Levy, Roger},
   year={2023},
   pages={5153–5176} }

@misc{pan2023logiclmempoweringlargelanguage,
      title={Logic-LM: Empowering Large Language Models with Symbolic Solvers for Faithful Logical Reasoning}, 
      author={Liangming Pan and Alon Albalak and Xinyi Wang and William Yang Wang},
      year={2023},
      eprint={2305.12295},
      archivePrefix={arXiv},
      primaryClass={cs.CL},
      url={https://arxiv.org/abs/2305.12295}, 
}

@misc{xu2024faithfullogicalreasoningsymbolic,
      title={Faithful Logical Reasoning via Symbolic Chain-of-Thought}, 
      author={Jundong Xu and Hao Fei and Liangming Pan and Qian Liu and Mong-Li Lee and Wynne Hsu},
      year={2024},
      eprint={2405.18357},
      archivePrefix={arXiv},
      primaryClass={cs.CL},
      url={https://arxiv.org/abs/2405.18357}, 
}

@misc{qwen2025qwen25technicalreport,
      title={Qwen2.5 Technical Report}, 
      author={Qwen and : and An Yang and Baosong Yang and Beichen Zhang and Binyuan Hui and Bo Zheng and Bowen Yu and Chengyuan Li and Dayiheng Liu and Fei Huang and Haoran Wei and Huan Lin and Jian Yang and Jianhong Tu and Jianwei Zhang and Jianxin Yang and Jiaxi Yang and Jingren Zhou and Junyang Lin and Kai Dang and Keming Lu and Keqin Bao and Kexin Yang and Le Yu and Mei Li and Mingfeng Xue and Pei Zhang and Qin Zhu and Rui Men and Runji Lin and Tianhao Li and Tianyi Tang and Tingyu Xia and Xingzhang Ren and Xuancheng Ren and Yang Fan and Yang Su and Yichang Zhang and Yu Wan and Yuqiong Liu and Zeyu Cui and Zhenru Zhang and Zihan Qiu},
      year={2025},
      eprint={2412.15115},
      archivePrefix={arXiv},
      primaryClass={cs.CL},
      url={https://arxiv.org/abs/2412.15115}, 
}

@misc{grattafiori2024llama3herdmodels,
      title={The Llama 3 Herd of Models}, 
      author={Aaron Grattafiori et al.},
      year={2024},
      eprint={2407.21783},
      archivePrefix={arXiv},
      primaryClass={cs.AI},
      url={https://arxiv.org/abs/2407.21783}, 
}

@misc{onoe2021creakdatasetcommonsensereasoning,
      title={CREAK: A Dataset for Commonsense Reasoning over Entity Knowledge}, 
      author={Yasumasa Onoe and Michael J. Q. Zhang and Eunsol Choi and Greg Durrett},
      year={2021},
      eprint={2109.01653},
      archivePrefix={arXiv},
      primaryClass={cs.CL},
      url={https://arxiv.org/abs/2109.01653}, 
}

@misc{zhang2024sac3reliablehallucinationdetection,
      title={SAC3: Reliable Hallucination Detection in Black-Box Language Models via Semantic-aware Cross-check Consistency}, 
      author={Jiaxin Zhang and Zhuohang Li and Kamalika Das and Bradley A. Malin and Sricharan Kumar},
      year={2024},
      eprint={2311.01740},
      archivePrefix={arXiv},
      primaryClass={cs.CL},
      url={https://arxiv.org/abs/2311.01740}, 
}

@inproceedings{thorne-etal-2018-fever,
    title = "{FEVER}: a Large-scale Dataset for Fact Extraction and {VER}ification",
    author = "Thorne, James  and
      Vlachos, Andreas  and
      Christodoulopoulos, Christos  and
      Mittal, Arpit",
    editor = "Walker, Marilyn  and
      Ji, Heng  and
      Stent, Amanda",
    booktitle = "Proceedings of the 2018 Conference of the North {A}merican Chapter of the Association for Computational Linguistics: Human Language Technologies, Volume 1 (Long Papers)",
    month = jun,
    year = "2018",
    address = "New Orleans, Louisiana",
    publisher = "Association for Computational Linguistics",
    url = "https://aclanthology.org/N18-1074/",
    doi = "10.18653/v1/N18-1074",
    pages = "809--819"
}

@article{Li_Ji_Wu_Li_Qin_Wei_Zimmermann_2024, 
title={Panoptic Scene Graph Generation with Semantics-Prototype Learning}, 
volume={38}, url={https://ojs.aaai.org/index.php/AAAI/article/view/28098}, 
DOI={10.1609/aaai.v38i4.28098}, 
number={4}, 
journal={Proceedings of the AAAI Conference on Artificial Intelligence}, 
author={Li, Li and Ji, Wei and Wu, Yiming and Li, Mengze and Qin, You and Wei, Lina and Zimmermann, Roger}, 
year={2024}, 
month={Mar.}, 
pages={3145-3153} }

@inproceedings{limm,
author = {Li, Li and Wang, Chenwei and Qin, You and Ji, Wei and Liang, Renjie},
title = {Biased-Predicate Annotation Identification via Unbiased Visual Predicate Representation},
year = {2023},
isbn = {9798400701085},
publisher = {Association for Computing Machinery},
address = {New York, NY, USA},
url = {https://doi.org/10.1145/3581783.3611847},
doi = {10.1145/3581783.3611847},
booktitle = {Proceedings of the 31st ACM International Conference on Multimedia},
pages = {4410–4420},
numpages = {11},
location = {Ottawa ON, Canada},
series = {MM '23}
}

@InProceedings{Li_2025_CVPR,
    author    = {Li, Shawn and Gong, Huixian and Dong, Hao and Yang, Tiankai and Tu, Zhengzhong and Zhao, Yue},
    title     = {DPU: Dynamic Prototype Updating for Multimodal Out-of-Distribution Detection},
    booktitle = {Proceedings of the IEEE/CVF Conference on Computer Vision and Pattern Recognition (CVPR)},
    month     = {June},
    year      = {2025},
    pages     = {10193-10202}
}

@InProceedings{li2025secureondevicevideoood,
    title={Secure On-Device Video OOD Detection Without Backpropagation}, 
    author={Shawn Li and Peilin Cai and Yuxiao Zhou and Zhiyu Ni and Renjie Liang and You Qin and Yi Nian and Zhengzhong Tu and Xiyang Hu and Yue Zhao},
    booktitle = {International Conference on Computer Vision (ICCV)},
    month     = {October},
    year      = {2025}
}

@inproceedings{li-etal-2025-treble,
    title = "Treble Counterfactual {VLM}s: A Causal Approach to Hallucination",
    author = "Shawn, Li  and
      Qu, Jiashu  and
      Song, Linxin  and
      Zhou, Yuxiao  and
      Qin, Yuehan  and
      Yang, Tiankai  and
      Zhao, Yue",
    booktitle = "Association for Computational Linguistics: EMNLP 2025",
    month = nov,
    year = "2025",
    address = "Suzhou, China",
    publisher = "Association for Computational Linguistics",
    pages = "18423--18434",
    ISBN = "979-8-89176-335-7",
}

@misc{li2025personalizedconversationalbenchmarksimulating,
      title={A Personalized Conversational Benchmark: Towards Simulating Personalized Conversations}, 
      author={Li Li and Peilin Cai and Ryan A. Rossi and Franck Dernoncourt and Branislav Kveton and Junda Wu and Tong Yu and Linxin Song and Tiankai Yang and Yuehan Qin and Nesreen K. Ahmed and Samyadeep Basu and Subhojyoti Mukherjee and Ruiyi Zhang and Zhengmian Hu and Bo Ni and Yuxiao Zhou and Zichao Wang and Yue Huang and Yu Wang and Xiangliang Zhang and Philip S. Yu and Xiyang Hu and Yue Zhao},
      year={2025},
      eprint={2505.14106},
      archivePrefix={arXiv},
      primaryClass={cs.CL},
      url={https://arxiv.org/abs/2505.14106}, 
}

@misc{li2026defensespromptattackslearn,
      title={Defenses Against Prompt Attacks Learn Surface Heuristics}, 
      author={Shawn Li and Chenxiao Yu and Zhiyu Ni and Hao Li and Charith Peris and Chaowei Xiao and Yue Zhao},
      year={2026},
      eprint={2601.07185},
      archivePrefix={arXiv},
      primaryClass={cs.CR},
      url={https://arxiv.org/abs/2601.07185}, 
}
\bibliographystyle{tmlr}

\newpage
\appendix
\section{Appendix}
\subsection{KG-FPQ Dataset Details}\label{sec:kgdataset}
In KoPL \citep{zhu2024kgfpqevaluatingfactualityhallucination}, each entity is linked to a specific concept, such as \textit{Leonardo da Vinci} being connected to the concept of an \textit{artist}. The knowledge graph includes 794 distinct concepts, categorized into domains based on general knowledge, enabling domain-based entity classification. For the art domain, the authors of \citep{zhu2024kgfpqevaluatingfactualityhallucination} manually selected 33 relations, ensuring that each relation is relevant to its domain and informative, avoiding ambiguity. For example, the relation \textit{artist} is linked to the Art domain, while \textit{family} is more ambiguous and excluded. Table~\ref{tab:dataset} shows the representative concepts, relations and subjects in the art domain of KG-FPQ.
The dataset comprises 4969 questions in the discriminative task for the art domain, with each true premise question modified using the following editing methods: Neighbor-Same-Concept (NSC), Neighbor-Different-Concept (NDC), Not-Neighbor-Same-Concept (NNSC), Not-Neighbor-Different-Concept (NNDC), Not-Neighbor-Same-Relation (NNSR), and Not-Neighbor-Different-Relation (NNDR).

\begin{table}[h!]
\centering
\resizebox{\columnwidth}{!}{%
\begin{tabular}{lllllll}
\hline
\textbf{Domain} & \textbf{Concept e.g.} & 
\textbf{Concept Qty} & \textbf{Subject e.g.} & \textbf{Subject Qty} & \textbf{Relation e.g.} & \textbf{Relation Qty} \\ 
\hline
 & film & & Titanic & & cast member & \\ 
Art & television series & 44 & Modern Family & 1754 & composer & 33 \\ 
 & drama & & Hamlet & & narrative location & \\ \hline
\end{tabular}%
}
\caption{Representative concepts, relations, and subjects in KG-FPQ art domain.}
\label{tab:dataset}
\end{table}

\subsection{CREAK Dataset Details}\label{sec:creakdataset}
The CREAK dataset \cite{onoe2021creakdatasetcommonsensereasoning}  is designed to test whether language models can combine factual knowledge about specific entities with commonsense reasoning. It consists of 13k English claims covering 2.7k entities, each labeled as true or false.  These claims require reasoning that bridges factual information (e.g., “Harry Potter is a wizard”) with unstated commonsense inferences (e.g., “If someone is good at a skill, they can teach it”). Unlike prior commonsense benchmarks that focus on generic physical or social scenarios, CREAK emphasizes entity-grounded reasoning and assesses whether models can verify claims that depend on both knowledge retrieval and implicit reasoning. Table~\ref{tab:creak-stats} summarizes the dataset statistics.

\begin{table}[h!]
\centering
\begin{tabular}{lcccccc}
\hline
\textbf{Split} & \multicolumn{3}{c}{\textbf{\# Claims}} & \textbf{Average Length} & \textbf{\# Unique Entities} & \textbf{Vocab Size} \\
\cline{2-4}
 & \textbf{Total} & \textbf{True} & \textbf{False} & (\# tokens) &  &  \\
\hline
Train & 10{,}176 & 5{,}088 & 5{,}088 & 10.8 & 2{,}096 & 19{,}006 \\
Dev & 1{,}371 & 691 & 680 & 9.7 & 531 & 4{,}520 \\
Test & 1{,}371 & 707 & 664 & 9.9 & 538 & 4{,}620 \\
Test (Contrast) & 500 & 250 & 250 & 10.0 & 226 & 1{,}596 \\
\hline
\end{tabular}
\caption{Data statistics of \textsc{CREAK}.}
\label{tab:creak-stats}
\end{table}

\subsection{Prompt Details}\label{sec:prompt}
The following prompt is used to combine the information retrieved from the knowledge graph $G$ (context) and the query logical form $\mathcal{L}(q)$ (query) to form the input to the LLMs discussed in the Section \textit{False Premise Detection with Logical Form}.

\texttt{Given the context below, does the following question contain a false premise? Answer with 'Yes' or 'No' only. Note that the context is provided
as valid facts in a triple.
Context: [context].
Query: [query].}

We use the following prompt for logical form conversion:

\texttt{You are given a question. The task is to: 1) define all the predicates used in the question. 2) parse the question into logic rules based on the defined predicates 3) translate any logical rules implied by the question. 4) convert the question into a logical form using predicate logic. Provide your final answer in the following format: Logical form: Predicate1(entity1, entity2). Keep all expressions concise and consistent. Use standard predicate logic notation.}

\subsection{Additional Results}\label{add_result}
\subsubsection{Query-level Hallucination Mitigation Analysis}\label{app:premise_breakdown}
\begin{table}[h]
\centering
\small
\setlength{\tabcolsep}{6pt}
\begin{tabular}{lccc}
\hline
\textbf{Model} & \textbf{Extra Correct} & \textbf{FPQ Improved} & \textbf{TPQ Change} \\
\hline
GPT-4o-mini & 427 & 392 & +35 \\
GPT-3.5     & 50  & 60  & -10 \\
Llama-3.1   & 139 & 115 & +24 \\
Mistral-7B  & 94  & 108 & -14 \\
Qwen2.5     & 139 & 124 & +15 \\
Qwen1.5     & 94  & 74  & +20 \\
\hline
\end{tabular}
\caption{\update{Premise-level breakdown of hallucination mitigation improvements.
\textit{Extra Correct} counts queries newly answered correctly compared to direct prompting.
\textit{FPQ Improved} counts false-premise queries corrected after false-premise detection.
\textit{TPQ Change} indicates net changes on true-premise queries after the detection and informing process.}}
\label{tab:premise_breakdown}
\end{table}

\subsubsection{Logical Form Ablation Study}
\label{app:logical_form_ablation}
\begin{table}[h]
\centering
\small
\setlength{\tabcolsep}{5pt}
\begin{tabular}{lcccc}
\hline
 & \textbf{Full} & \textbf{w/o Rel.} & \textbf{w/o Ent$_1$} & \textbf{w/o Ent$_2$} \\
\hline
\multicolumn{5}{c}{Logical Form for Retrieval + Original Query for Detection} \\
\hline
TPR & 0.377 & 0.044 & 0.067 & 0.011   \\
TNR & 0.866 & 0.000 & 0.067 & 0.267   \\
FPR & 0.133 & 1.000 & 0.933 & 0.733   \\
FNR & 0.622 & 0.956 & 0.933 & 0.989   \\
F1  & 0.540 0.073 & 0.109 & 0.020   \\
Acc & 0.800 & 0.038 & 0.067 & 0.048   \\
\hline
\multicolumn{5}{c}{Logical Form for Both Stages} \\
\hline
TPR & 0.600 & 0.044 & 0.067 & 0.044   \\
TNR & 0.866 & 0.000 & 0.133 & 0.267   \\
FPR & 0.133 & 1.000 & 0.867 & 0.733   \\
FNR & 0.400 & 0.956 & 0.933 & 0.956   \\
F1  & 0.740 & 0.073 & 0.109 & 0.076   \\
Acc & 0.829 & 0.038 & 0.076 & 0.076   \\
\hline
\end{tabular}
\caption{Logical form ablation results using \textbf{with RAG}.}
\label{tab:lf_ablation_rag}
\end{table}

\begin{table}[h]
\centering
\small
\setlength{\tabcolsep}{5pt}
\begin{tabular}{lcccc}
\hline
 & \textbf{Full} & \textbf{w/o Rel.} & \textbf{w/o Ent$_1$} & \textbf{w/o Ent$_2$} \\
\hline
\multicolumn{5}{c}{Logical Form for Retrieval + Original Query for Detection} \\
\hline
TPR & 0.822 & 0.178 & 0.078 & 0.111   \\
TNR & 0.933 & 0.133 & 0.067 & 0.133   \\
FPR & 0.066 & 0.867 & 0.933 & 0.867   \\
FNR & 0.177 & 0.822 & 0.922 & 0.889   \\
F1  & 0.869 & 0.270 & 0.124 & 0.177   \\
Acc & 0.838 & 0.171 & 0.076 & 0.114   \\
\hline
\multicolumn{5}{c}{Logical Form for Both Stages} \\
\hline
TPR & 0.944 & 0.167 & 0.089 & 0.133   \\
TNR & 0.991 & 0.133 & 0.133 & 0.200   \\
FPR & 0.009 & 0.867 & 0.867 & 0.800   \\
FNR & 0.056 & 0.833 & 0.911 & 0.867   \\
F1  & 0.971 & 0.254 & 0.144 & 0.211   \\
Acc & 0.952 & 0.162 & 0.095 & 0.143   \\
\hline
\end{tabular}
\caption{Logical form ablation results using \textbf{G-retriever}.}
\label{tab:lf_ablation_gretriever}
\end{table}

\begin{table}[h]
\centering
\small
\setlength{\tabcolsep}{5pt}
\begin{tabular}{lcccc}
\hline
 & \textbf{Full} & \textbf{w/o Rel.} & \textbf{w/o Ent$_1$} & \textbf{w/o Ent$_2$} \\
\hline
\multicolumn{5}{c}{Logical Form for Retrieval + Original Query for Detection} \\
\hline
TPR & 0.089 & 0.056 & 0.089 & 0.078   \\
TNR & 0.933 & 0.133 & 0.200 & 0.267   \\
FPR & 0.067 & 0.867 & 0.800 & 0.733   \\
FNR & 0.911 & 0.944 & 0.911 & 0.922   \\
F1  & 0.162 & 0.093 & 0.145 & 0.130   \\
Acc & 0.813 & 0.067 & 0.105 & 0.105   \\
\hline
\multicolumn{5}{c}{Logical Form for Both Stages} \\
\hline
TPR & 0.089 & 0.100 & 0.100 & 0.067   \\
TNR & 0.933 & 0.133 & 0.200 & 0.267   \\
FPR & 0.067 & 0.867 & 0.800 & 0.733   \\
FNR & 0.911 & 0.900 & 0.900 & 0.933   \\
F1  & 0.162 & 0.163 & 0.162 & 0.112   \\
Acc & 0.813 & 0.104 & 0.114 & 0.095   \\
\hline
\end{tabular}
\caption{Logical form ablation results using \textbf{GraphRAG/ToG}.}
\label{tab:lf_ablation_graphrag}
\end{table}
\update{We examine the contribution of individual components in the logical form, such as entity arguments and relational structure, by analyzing how the removal of specific elements affects false-premise detection across different retrieval backends.
We consider variants where the logical form is used for retrieval while the original query is used for false-premise detection, as well as variants where the logical form is applied to both stages.
The results are shown in Tab. \ref{tab:lf_ablation_rag}, \ref{tab:lf_ablation_gretriever}, \ref{tab:lf_ablation_graphrag}. Across all retrieval backends, removing relational structure or entity arguments from the logical form consistently degrades performance. This suggests that the logical form is most effective when its constituent elements are jointly preserved, and that each component contributes complementary information for identifying unsupported premises.}

\subsubsection{Comparison with Post-hoc Method}\label{sec:add_posthoc}
Tab.~\ref{tab:contrastive_vs_ours} presents a performance comparison between Contrastive Decoding \citep{shi2023trustingevidencehallucinatecontextaware}, a post-hoc hallucination mitigation method, and other LLMs (Mistral-7B, Qwen1.5, Qwen2.5-7B-Instruct, Llama-3.1-8B-Instruct). Our method achieves improved performance over Contrastive Decoding on all models except Mistral-7B.

\subsubsection{False Premise Detection}\label{sec:llama_detection}
\update{We additionally evaluate Llama-3.1-8b}, as well as GPT-3.5-turbo and G-retriever on the false premise detection task using our method. The results are presented below (Tab.~\ref{tab:detection_llama} and Tab.~\ref{tab:1}, Fig.~\ref{fig:radar}). Notably, when original queries are used in either retrieval, false premise detection, or both stages,
despite achieving high accuracy (91.11\%), G-retriever shows a markedly lower TPR (37.78\%) compared to the first configuration. This suggests that relying on original queries alone, or in combination with logical forms in only one stage for detection, can achieve high accuracy due to correctly identifying negatives, it is less effective at capturing false premises, which is the primary focus of our task.

\begin{table}[t]
\centering
\small
\setlength{\tabcolsep}{6pt}
\begin{tabular}{lcccc}
\hline
\textbf{Metric} & \textbf{Direct Claim} & \textbf{with RAG} & \textbf{G-retriever} & \textbf{GraphRAG/ToG} \\
\hline
\multicolumn{5}{c}{Original Query for Both Stages} \\
\hline
TPR      & 0.878 & 0.811 & 0.644 & 0.800 \\
TNR      & 0.200 & 0.467 & 0.467 & 0.267 \\
FPR      & 0.800 & 0.533 & 0.533 & 0.733 \\
FNR      & 0.122 & 0.189 & 0.356 & 0.200 \\
F1       & 0.873 & 0.854 & 0.744 & 0.832 \\
Accuracy & 0.781 & 0.762 & 0.619 & 0.724 \\
\hline
\multicolumn{5}{c}{Logical Form for Retrieval and Original Query for False Premise Detection} \\
\hline
TPR      & 0.878 & 0.811 & 0.711 & 0.800 \\
TNR      & 0.200 & 0.400 & 0.333 & 0.267 \\
FPR      & 0.800 & 0.600 & 0.667 & 0.733 \\
FNR      & 0.122 & 0.189 & 0.289 & 0.200 \\
F1       & 0.873 & 0.849 & 0.780 & 0.832 \\
Accuracy & 0.781 & 0.752 & 0.657 & 0.724 \\
\hline
\multicolumn{5}{c}{Logical Form for Both Stages} \\
\hline
TPR      & 0.878 & 0.800 & 0.811 & 0.800 \\
TNR      & 0.200 & 0.867 & 0.400 & 0.267 \\
FPR      & 0.800 & 0.133 & 0.600 & 0.733 \\
FNR      & 0.122 & 0.200 & 0.189 & 0.200 \\
F1       & 0.873 & 0.878 & 0.849 & 0.832 \\
Accuracy & 0.781 & 0.810 & 0.752 & 0.724 \\
\hline
\end{tabular}
\caption{\update{KG-FPQ dataset: comparison of performance metrics across different retrieval methods using original queries and logical forms at different stages using Llama-3.1-8b.}}
\label{tab:detection_llama}
\end{table}

\subsubsection{Premise Detection Baseline}\label{sec:sac3}
We include SAC3 \cite{zhang2024sac3reliablehallucinationdetection} as baseline for premise detection for the KG-FPQ dataset. Our proposed approach achieves better performance when considering both F1 score and Accuracy (see Table~\ref{tab:false_premise_detection_perf} for comparison).
\begin{table}[h]
\centering
\begin{tabular}{lcccccc}
\hline
\textbf{Method} & \textbf{TPR} & \textbf{TNR} & \textbf{FPR} & \textbf{FNR} & \textbf{F1} & \textbf{Acc} \\
\hline
SAC3 & 81.1 & 73.3 & 26.7 & 18.9 & 87.4 & 80.0 \\
\hline
\end{tabular}
\caption{Performance of SAC3 on premise detection on KG-FPQ dataset.}
\label{tab:sac3_results}
\end{table}

\begin{table}[h]
\centering
\renewcommand{\arraystretch}{1.1}
\begin{tabular}{lc}
\hline
\textbf{Metric} & \textbf{G-retriever} \\
\hline
\multicolumn{2}{c}{Original Query for Both Stages} \\
\hline
True Positives (TP\%)  & 37.78  \\
True Negatives (TN\%)  & 100.00 \\
False Positives (FP\%) & 0.00   \\
False Negatives (FN\%) & 62.22  \\
F1 Score (\%)          & 54.84  \\
Accuracy (\%)          & 91.11  \\
\hline
\multicolumn{2}{c}{Logical Form + Original Query} \\
\hline
True Positives (TP\%)  & 37.78  \\
True Negatives (TN\%)  & 100.00 \\
False Positives (FP\%) & 0.00   \\
False Negatives (FN\%) & 62.22  \\
F1 Score (\%)          & 54.84  \\
Accuracy (\%)          & 91.11  \\
\hline
\multicolumn{2}{c}{Logical Form for Both Stages} \\
\hline
True Positives (TP\%)  & 75.56  \\
True Negatives (TN\%)  & 80.00  \\
False Positives (FP\%) & 20.00  \\
False Negatives (FN\%) & 24.44  \\
F1 Score (\%)          & 84.47  \\
Accuracy (\%)          & 79.37  \\
\hline
\end{tabular}
\caption{False Premise Detection Performance using GPT-3.5-turbo and G-retriever.}
\label{tab:1}
\end{table}

\begin{figure*}[htbp]
    \centering
    \includegraphics[width=\linewidth]{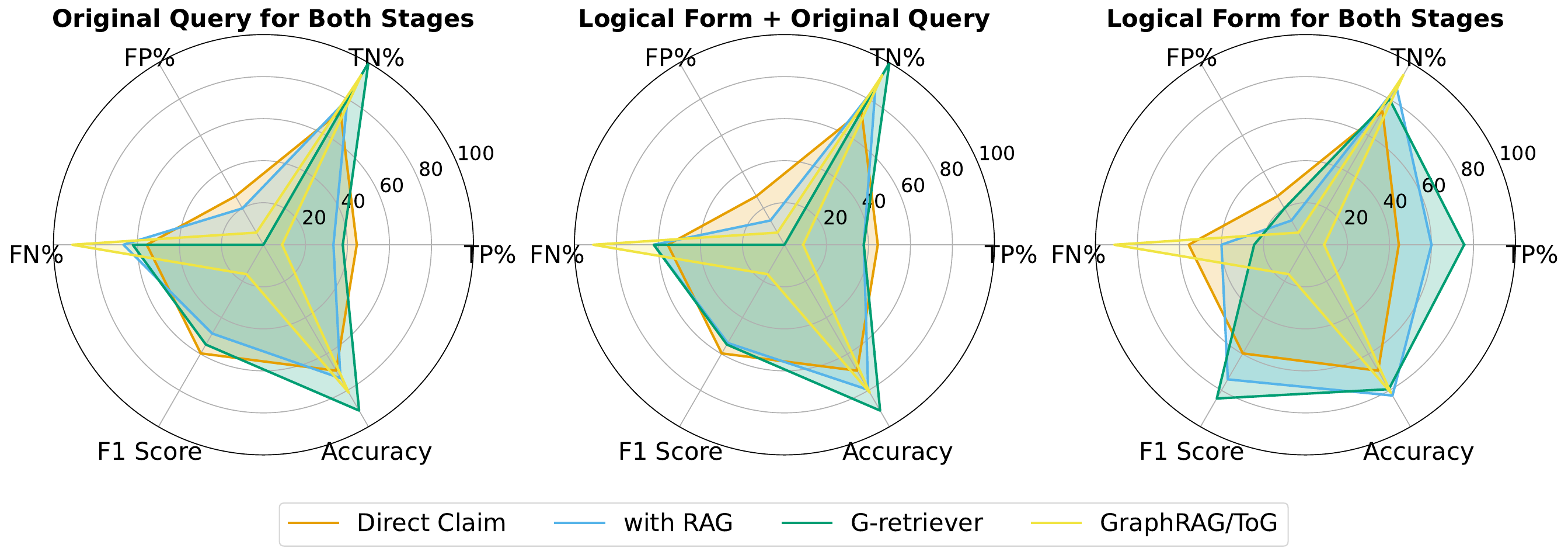}
    \caption{Additional comparison of performance metrics across different retrieval methods using logical forms and/or original queries.}
    \label{fig:radar3}
\end{figure*}

\begin{table*}[h]
\centering
\begin{tabular}{lcccc}
\hline
& \textbf{Mistral-7B} & \textbf{Qwen1.5} & \textbf{Qwen2.5-7B-Instruct} & \textbf{Llama-3.1-8B-Instruct} \\
\hline
Contrastive Decoding & 89.5 & 76.2 & 85.7 & 84.8 \\
Ours & 87.6 & 89.5 & 92.4 & 86.7 \\
\hline
\end{tabular}
\caption{Comparison between contrastive decoding and our method across different LLMs. Note: GPT-3.5 and GPT-4o-mini are not included as logits are not available for contrastive decoding approach.}
\label{tab:contrastive_vs_ours}
\end{table*}

\subsubsection{Case Study}\label{sec:case_study}
We perform a case study demonstrating how our method enhances multi-hop question performance. We use the \textit{with RAG} method example here, where the edited object is not directly connected to the subject within the KoPL knowledge graph.

\textbf{Question}: \texttt{Is "The Dark Knight" the recipient of the 16th Screen Actors Guild Awards?}

\textbf{Answer}: \texttt{Yes, "The Dark Knight" won the award for Outstanding Performance by a Stunt Ensemble in a Motion Picture at the 16th Screen Actors Guild Awards.}

\textbf{Logical Form}: 
\texttt{
is a recipient of('The Dark Knight', 16th Screen Actors Guild Awards)
}

\textbf{Retrieved Graph Triple}: 
\texttt{
['The Dark Knight', 'award received', '81st Academy Awards']
}

After Detecting and Informing LLM of the Presence of a False Premise:

\textbf{Corrected Answer}: 

\texttt{No, "The Dark Knight" was not the recipient of the 16th Screen Actors Guild Awards. That year’s SAG Award for Outstanding Performance by a Cast in a Motion Picture went to Inglourious Basterds, not The Dark Knight.}

\subsection{\update{Evaluation under Longer Queries}}
\label{app:long_context}

\update{Longer questions often introduce additional contextual cues, compositional structures, and distracting details, which can obscure the core false assumption and make premise verification more challenging. Evaluating under such settings is important for understanding whether a model can robustly identify and reason about incorrect premises rather than relying on surface-level patterns.}

\update{As an initial step, we select the first 105 questions from the KG-FPQ dataset and use an LLM (gpt-4o-mini) to rewrite them into longer, more natural user queries while preserving the original (true or false) premises. We then evaluate our method on both false-premise detection and hallucination mitigation under this longer-context setting. The prompt used for query rewriting, along with the rewritten queries, is released in accompanying repository\footnote{\url{https://github.com/yqin43/premise-verification}}. The False Premise Detection result and Hallucination Mitigation result are shown in Tab.~\ref{tab:long_fpq_detection} and Tab.~\ref{tab:long_hallu}. Overall, the results on this longer query subset show that the proposed method consistently improves performances on both false-premise detection and hallucination mitigation tasks, suggesting that its effectiveness extends beyond short or minimally phrased queries.}

\begin{table}[h]
\centering
\small
\setlength{\tabcolsep}{6pt}
\begin{tabular}{lcccc}
\hline
\textbf{Metric} & \textbf{Direct Claim} & \textbf{with RAG} & \textbf{G-retriever} & \textbf{GraphRAG/ToG} \\
\hline
\multicolumn{5}{c}{\textit{Original Query for Both Stages}} \\
\hline
TPR & 2.22 & 85.56 & 68.89 & 92.22 \\
TNR & 50.00 & 93.33 & 73.33 & 53.33 \\
FPR & 50.00 & 6.67 & 26.67 & 46.67 \\
FNR & 97.78 & 14.44 & 31.11 & 7.78 \\
F1  & 4.04 & 91.67 & 79.49 & 92.22 \\
Accuracy & 8.65 & 86.67 & 69.52 & 86.67 \\
\hline
\multicolumn{5}{c}{\textit{Logical Form for Retrieval and Original Query for Detection}} \\
\hline
TPR & 2.22 & 94.44 & 93.33 & 92.22 \\
TNR & 50.00 & 73.33 & 53.33 & 53.33 \\
FPR & 50.00 & 26.67 & 46.67 & 46.67 \\
FNR & 97.78 & 5.56 & 6.67 & 7.78 \\
F1  & 4.04 & 94.97 & 92.82 & 92.22 \\
Accuracy & 8.65 & 91.43 & 87.62 & 86.67 \\
\hline
\multicolumn{5}{c}{\textit{Logical Form for Both Stages}} \\
\hline
TPR & 2.22 & 94.44 & 97.78 & 92.22 \\
TNR & 50.00 & 73.33 & 46.67 & 53.33 \\
FPR & 50.00 & 26.67 & 53.33 & 46.67 \\
FNR & 97.78 & 5.56 & 2.22 & 7.78 \\
F1  & 4.04 & 94.97 & 94.62 & 92.22 \\
Accuracy & 8.65 & 91.43 & 90.48 & 86.67 \\
\hline
\end{tabular}
\caption{\update{False-premise detection performance under longer queries.}}
\label{tab:long_fpq_detection}
\end{table}

\begin{table}[h]
\centering
\small
\setlength{\tabcolsep}{8pt}
\begin{tabular}{lc}
\hline
\textbf{Setting} & \textbf{Accuracy} \\
\hline
Direct Ask & 0.933 \\
Ours       & 0.952 \\
\hline
\end{tabular}
\caption{\update{Hallucination mitigation accuracy under longer queries using GPT-4o-mini.}}
\label{tab:long_hallu}
\end{table}

\subsection{Logical Form Correctness Validation}\label{sec:annotator}
\update{Our current human evaluation was conducted by two Ph.D. students with relevant NLP/LLM research experience. All annotators reached full agreement, resulting in an inter-annotator agreement of 1.0}.

\subsection{Additional Experiment Setup}
All models are implemented and run on a multi-NVIDIA RTX 6000 Ada workstation. For \textit{Logical Form Extraction} and \textit{Retrieval}, we set parameters temperature $= 0$ and top\_p = 1.

\end{document}